\documentclass{article}

\usepackage{microtype}
\usepackage{graphicx}
\usepackage{subcaption}
\usepackage{booktabs} 
\usepackage[table]{xcolor}
\usepackage[pagebackref,breaklinks,colorlinks,citecolor=cvprblue]{hyperref}
\usepackage{graphicx}
\usepackage{float,epstopdf}
\usepackage{bbm}

\usepackage{microtype}

\usepackage{subcaption}
\usepackage{booktabs}

\usepackage{amsmath}
\usepackage{amssymb}
\usepackage{mathtools}
\usepackage{amsthm}
\usepackage{dsfont}
\usepackage{multicol}
\usepackage{multirow} 
\usepackage{amsfonts} 
\usepackage{mathrsfs}
\usepackage{fancyhdr}
\usepackage[amssymb, thickqspace]{SIunits}
\usepackage{enumitem}
\usepackage{pgfplotstable}
\usepackage{arydshln}
\usepackage{lipsum}		
\usepackage{hyperref}

\usepackage{cases}

\usepackage[ruled, linesnumbered]{algorithm2e}

\usepackage{url}

\usepackage{thmtools}
\usepackage{thm-restate}
\usepackage{tabu}
\makeatletter
\def\munderbar#1{\underline{\sbox\tw@{$#1$}\dp\tw@\z@\box\tw@}}
\makeatother

\newtheorem{definition}{Definition}[section]
\newtheorem{theorem}[definition]{Theorem}
\newtheorem{lemma}[definition]{Lemma}

\newtheorem{proposition}[definition]{Proposition}

\newtheorem{axiom}{Axiom}
\newtheorem{property}[axiom]{Property}

\AddToHook{cmd/appendix/before}{%
  \setcounter{axiom}{0}%
}

\newcommand{\blue}[1]{{\color{blue}#1}}

\newcommand{\Ic}{\mc I}

\newcommand{\Sh}{\phi}
\newcommand{\My}{\psi}

\newcommand{\ShI}{\Phi}
\newcommand{\MyI}{\Psi}

\newcommand{\be}{\begin{equation}}
\newcommand{\ee}{\end{equation}}
\newcommand{\bea}{\begin{equation*}\begin{aligned}}
\newcommand{\eea}{\end{aligned}\end{equation*}}

\newcommand{\R}{\mathbb{R}}

\newcommand{\bL}{\textbf{L}}
\newcommand{\bD}{\textbf{D}}
\newcommand{\bS}{\textbf{S}}
\newcommand{\bE}{\textbf{E}}
\newcommand{\bID}{\textbf{ID}}

\newcommand{\bCE}{\textbf{CE}}
\newcommand{\bCF}{\textbf{CF}}

\newcommand{\bRNP}{\textbf{RNP}}

\newcommand{\mc}{\mathcal}

\newcommand{\ie}{{\it i.e.}}

\DeclareMathOperator{\st}{s.t.}

\newcommand{\ve}[1]{\mathbf{#1}}
\usepackage{pgfplotstable}
\usepackage{hyperref}

\usepackage[accepted]{icml2024}

\usepackage[textsize=tiny]{todonotes}

\icmltitlerunning{Explaining Graph Neural Networks via Structure-aware Interaction Index}

\begin{document}

\twocolumn[
\icmltitle{Explaining Graph Neural Networks via Structure-aware Interaction Index}




\begin{icmlauthorlist}
\icmlauthor{Ngoc Bui}{yale}
\icmlauthor{Hieu Trung Nguyen}{vinai}
\icmlauthor{Viet Anh Nguyen}{cuhk}
\icmlauthor{Rex Ying}{yale}
\end{icmlauthorlist}

\icmlaffiliation{yale}{Yale University}
\icmlaffiliation{vinai}{VinAI Research}
\icmlaffiliation{cuhk}{The Chinese University of Hong Kong}

\icmlcorrespondingauthor{Ngoc Bui}{ngoc.bui@yale.edu}

\icmlkeywords{Graph Neural Networks, Explainablity, Cooperative Game Theory, Deep Learning}

\vskip 0.3in
]



\printAffiliationsAndNotice{}  

\begin{abstract}
The Shapley value is a prominent tool for interpreting black-box machine learning models thanks to its strong theoretical foundation. However, for models with structured inputs, such as graph neural networks, existing Shapley-based explainability approaches either focus solely on node-wise importance or neglect the graph structure when perturbing the input instance. This paper introduces the Myerson-Taylor interaction index that internalizes the graph structure into attributing the node values and the interaction values among nodes. Unlike the Shapley-based methods, the Myerson-Taylor index decomposes coalitions into components satisfying a pre-chosen connectivity criterion. We prove that the Myerson-Taylor index is the unique one that satisfies a system of five natural axioms accounting for graph structure and high-order interaction among nodes. Leveraging these properties, we propose Myerson-Taylor Structure-Aware Graph Explainer (MAGE), a novel explainer that uses the second-order Myerson-Taylor index to identify the most important motifs influencing the model prediction, both positively and negatively. Extensive experiments on various graph datasets and models demonstrate that our method consistently provides superior subgraph explanations compared to state-of-the-art methods. 
\end{abstract}

\section{Introduction}
Graph Neural Networks (GNNs) are ubiquitous thanks to their predictive power in many applications~\citep{zhou2020graph, wu2020comprehensive}. GNNs proliferate in various real-world applications, from natural language processing~\citep{wu2023graph}, image recognition and detection~\citep{han2022vision},  point cloud analysis~\citep{shi2020point, zhang2022not} to AI for science~\citep{sun2020graph}.  However, understanding the rationale behind the prediction of GNNs remains challenging. As GNNs are gaining popularity in high-stake domains, their (lack of) explainability becomes a growing concern~\citep{yuan2022explainability, kakkad2023survey}. 
Attempts towards explaining GNNs can be categorized into two main directions: \textit{white-} and \textit{black-box} explanability. White-box explainers~\citep{pope2019explainability, feng2022degree} typically necessitate access to a model's internal structures or gradients. In contrast, black-box explainers~\citep{ying2019gnnexplainer, luo2020parameterized, vu2020pgm, schlichtkrull2020interpreting, yuan2020xgnn} only require querying the model's output; hence they are more versatile and applicable to a broader range of architectures. 


Shapley value~\citep{shapley1953value} is a game theory concept successfully applied to explain black-box ML models in various domains, such as tabular or image data~\citep{lundberg2017unified}. However, adopting the Shapley values to models with graph input poses a significant challenge, mainly due to the combinatorial nature of graph structures. Several works~\citep{duval2021graphsvx, yuan2021explainability, ye2023same} have utilized the Shapley value to determine node importance in the graph input by perturbing the graph and measuring the change in the model's prediction when specific nodes are removed or ablated. The importance scores, or attribution scores, are used to identify a subset of nodes most influential to the model prediction~\citep{zhang2022gstarx}. However, existing approaches to leveraging Shapley values for graph data encounter several challenges.

\textit{First}, the Shapley value does not consider the graph structure when perturbing the graph input. This can lead to perturbed graphs that may be disconnected or pathological, which the GNN model does not observe during the training. Assessing the model on these pathological graphs may inject bias, adversely affecting the estimated attribution scores. Therefore, Shapley's attribution may not reflect the true importance of nodes~\citep{zhang2022gstarx}. 

\textit{Second}, most existing methods focus on attributing importance scores for nodes or edges individually~\cite{zhang2022gstarx, duval2021graphsvx}. They then apply a greedy algorithm to highlight a group of nodes/edges with the highest total node-wise importance. However, this approach is not suitable for applications requiring the identification of multiple motifs because the sum of node-wise importance is not sufficient to capture interaction among nodes within each motif~\citep{sundararajan2020shapley, masoomi2020instance, zhang2021interpreting}. Therefore, the highlighted nodes might be disconnected and unintuitive to humans.

\textit{Finally}, existing graph explainers only consider identifying substructures that positively affect the prediction and neglect the structures/motifs that may negatively affect the model prediction, hindering the model from giving a higher confidence score. Meanwhile, identifying negative structures can provide counterfactual reasoning, which can help practitioners avoid drawing misleading conclusions from the model's output. 

%

\noindent\textbf{Proposed work.} To address the aforementioned challenges, we introduce the Myerson-Taylor interaction index, which generalizes the Myerson values~\citep{myerson1977graphs} and Shapley-Taylor index~\citep{sundararajan2020shapley} to capture both structure information and high-order node interactions in the graph input when assigning importance scores. The Myerson-Taylor index incorporates the structure information into the Shapley values by only allowing interactions from connected nodes, thus potentially mitigating the Out-Of-Distribution (OOD) bias of the Shapley values.

Building on this, we propose a Myerson-Taylor Structure-aware Graph Explainer (MAGE) that leverages the second-order Myerson-Taylor index to compute pair-wise interaction among nodes. These pair-wise importances are used to compute the group attribution score, accounting for the importance of a subgraph to the model prediction. MAGE then solves an optimization model to find \textit{multiple} explanatory substructures that maximize the total absolute attribution score. Thus, MAGE can effectively identify both \textit{positive} and \textit{negative} motifs that contribute to the GNN output.

Extensive experiments on ten datasets and three GNN models to show MAGE's effectiveness in explaining GNN predictions. MAGE empirically outperforms seven popular, state-of-the-art baselines across diverse tasks, including molecular prediction, image, and sentiment classification. Specifically, we achieve up to 27.55\% increase in the explanation accuracy compared to the best baseline. 


\section{Related Work}
\textbf{Explainability for GNN models.} There are two main approaches to finding explanations for GNN models: \textit{self-interpretable methods} and \textit{post-hoc methods}~\citep{chen2023generative,kakkad2023survey, yuan2022explainability,amara2022graphframex}. Self-interpretable methods focus on designing model architectures that inherently generate explanations from input subgraphs~\citep{feng2022kergnns, miao2022interpretable}. In contrast, post-hoc explanation aims to construct explanations for existing trained models. Methods in post-hoc explainability for graph models can also be divided into two categories, \textit{black-box} and \textit{white-box}, depending on how they access the model information. White-box explainers usually require access to the internal structure, parameters, or gradients of the model~\citep{feng2022degree, schnake2021higher, baldassarre2019explainability, pope2019explainability, huang2024factorized}. Black-box explainers only require to query the model output to train a surrogate model~\citep{huang2022graphlime, zhang2021relex, pereira2023distill}, or generative model~\citep{chen2024d4explainer, wang2022gnninterpreter, chen2024tempme, shan2021reinforcement, yuan2020xgnn, lin2021generative, li2023dag} to construct explanations. Another black-box approach is perturbation-based methods~\citep{ying2019gnnexplainer, schlichtkrull2020interpreting, luo2020parameterized, wang2021towards, funke2022z, huang2024factorized}, which attribute node/edge importance by perturbing the input graphs and assess the change
in model's prediction. Specifically, within this domain, \citep{duval2021graphsvx, yuan2021explainability, ye2023same} propose to treat a subgraph as a supernode and other nodes of the graph as singletons. They then use the Shapley values of the supernode as its importance score. Although they leverage the graph input to compute Shapley values for $L$-hop neighbors around the supernode to reduce complexity, the underlying attribution score still relies on the Shapley value, which neglects the structural information~\citep{zhang2022gstarx}. To address this, \citet{zhang2022gstarx} propose to use Hamiache and Navarro (HN) value~\citep{hamiache2020associated} to incorporate graph structure by assigning zero weight for disconnected subgraphs. However, they only focus on node-wise importance and neglect node interactions when forming multiple motif groups.



\noindent\textbf{Cooperative game theory.} In ML's explainability with cooperative game theory,~\citet{grabisch1999axiomatic, sundararajan2020shapley, tsai2023faith} propose allocation rules to analyze high-order interactions among input features of ML models. \citet{zhang2021interpreting, masoomi2020instance}  study the group attribution, where features form non-separable coalitions, acting as unified groups. However, these works neglect graph inputs, thus omitting structural information in allocating importance scores. Recently,~\citet{zhang2022gstarx, homberg2023interpreting} adopted the Myerson value and HN-value to explain models with graph inputs. However, they only focus on node-wise importance. In this work, we propose a generalized allocation rule that considers both graph structure and high-order node interactions.

\section{Preliminaries} \label{sec:prel}

A graph input is denoted by $G = (V, E)$, where $V$ is the set of nodes, and $E$ is the set of edges. While a graph input may contain node and edge feature vectors, our framework does not exploit this information; hence, we drop the node and edge features from the graph notation. A black-box graph neural network (GNN) is represented by a function $f$ that takes $G$ as input, and it outputs the probability or logit value for predicting $G$ to be in a specific class. To simplify the notation, we use $f(V)$ to denote the output value: this notation highlights the node composition of the input, and the edges are taken implicitly. Similarly, for any subset of nodes $T \subseteq V$, we use $f(T)$ to denote the GNN output to the graph $(T, E_T)$, where $E_T$ is the collection of edges induced by $E$ with both endpoints in the subset $T$. We refer to $T$ as a subset of nodes or a subgraph interchangeably. To simplify notations, for any set $T$ and nodes $i$ and $j$, we use $T \cup i$ as a shorthand for $T \cup  \{i\}$, and $T \cup ij$ for $T \cup  \{i, j\}$. 


Given a set of nodes\footnote{To adapt to the graph explanation task, we use the terminologies `node' and `subset' throughout. In the game theory literature, `node' is called `player', and `subset' is called `coalition'.} $V$ and a function $f$, an attribution rule distributes the output value $f(V)$ to the members of $V$. 

\textbf{The Shapley value} quantifies the potential change in the model's prediction resulting from removing or ablating a particular node. The value attributed to each node is calculated as the average of its marginal contribution over all possible coalitions it could join.
\begin{definition}[Shapley value~\citep{shapley1953value}] \label{def:sh} Given a function $f$ and a set of nodes $V$, the Shapley value of node $i \in V$ is defined as
\[
    \phi_i = \frac{1}{|V|} \sum_{T \subseteq V \setminus i} \frac{1}{{|V| - 1 \choose |T|}} (f(T \cup i) - f(T)).
\]
\end{definition}
There are several drawbacks of the Shapley values: (i) it focuses solely on node-wise importance, thus failing to illustrate the interactions among nodes; (ii) it is not suitable for graph inputs because it disregards the graph connectivity structure. We next discuss several extensions in the literature that attempt to alleviate these drawbacks.

\textbf{The Shapley-Taylor index} aims to capture the interactions between nodes in the explanation task. To do this, \citet{sundararajan2020shapley} generalized the Shapley value to $k$-order explanation that attributes the model's prediction to interactions of subsets of nodes of size up to $k$. Denote $\delta_S(T)$ as the cooperative contribution (in terms of model output) of a subset of nodes $S$ when joining another subset $T$. Specifically, we can write $\delta_S(T)$ as
\be \label{eq:delta}
    \delta_S f(T) = \sum_{W \subseteq S} (-1)^{|S| - |W|} f(W \cup T).
\ee
Consider when $S = \{i\}$, then $\delta_i f(T) = f(T \cup i) - f(T)$, which is equivalent to the marginal contribution of $i$ to subset $T$. If $S = \{i, j\}$ with $i \neq j$, then 
\[
\delta_{ij} f(T) = f(T \cup ij) - f(T \cup i) - f(T \cup j) + f(T),
\] 
which is surplus created from interaction between to $i$ and $j$ when both joins a subset $T$. \citet{sundararajan2020shapley} defined the Shapley-Taylor index as follows.
\begin{definition}[Shapley-Taylor index~\citep{sundararajan2020shapley}]\label{def:sht}
    Given a function $f$ and a set of nodes $V$, the $k$-order Shapley-Taylor index of a subset $S \subseteq V$, $|S| \le k$ is defined as follows
    \[
        \ShI^k_S = \begin{dcases}
            \delta_S f(\emptyset) & \text{if}~~|S| < k, \\
            \frac{k}{|V|} \sum_{T \subseteq V \setminus S} \frac{1}{{|V|-1 \choose |T|}} \delta_S f(T) & \text{if}~~ |S| = k.
        \end{dcases}
    \]
\end{definition}
We observe the resemblance between the Shapley value in Def.~\ref{def:sh} and the Shapley-Taylor index in Def.~\ref{def:sht}: the branch $|S| = k$ of $\Phi^k_S$ has the same form with the Shapley value, except that $\Phi^k_S$ utilizes the difference function $\delta_S$ to capture the case when $S$ is not a singleton. Further, when $k = 1$, the Shapley-Taylor index recovers the Shapley value.

\begin{figure*}
    \centering
    \includegraphics[width=0.9\linewidth]{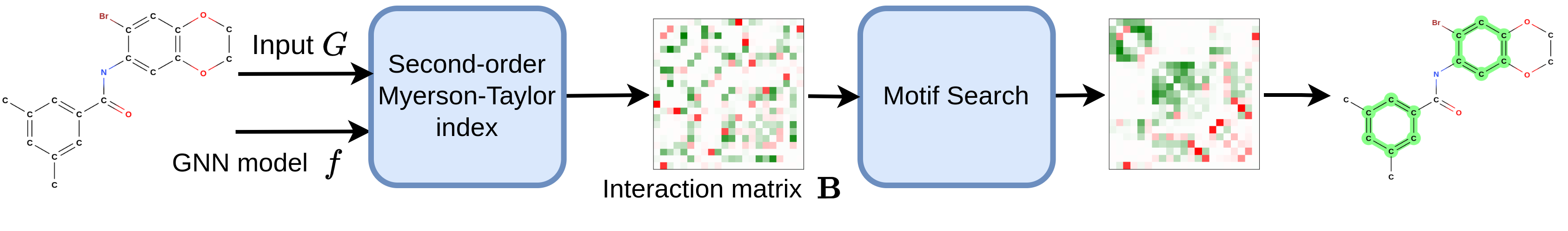}
    \vspace{-3mm}
    \caption{MAGE operates in two distinct phases: First, MAGE employs the second-order Myerson-Taylor index to calculate pairwise interactions among graph nodes, represented by an interaction matrix $\ve B$. This matrix $\ve B$ serves as the input for the motif optimization. This optimization module searches for the $m$ most influential motifs contributing to the model's prediction.}
    \label{fig:sage}
\end{figure*}


\textbf{The Myerson value} extends the Shapley value to account for interaction restrictions in graph settings~\citep{myerson1977graphs}. To formally delineate the Myerson value, let $\zeta(T)$ denote the set of connected components of the subgraph induced by $T \subseteq V$ in the graph $G = (V, E)$. We then define the interaction-restricted function as
\[
    f|_E(T) = \sum_{R \in \zeta(T)} f(R).
\]
If $T$ is a connected subgraph on $G$, thus $f|_E(T) = f(T)$. If $T$ is disconnected, the worth of subgraph $T$ is computed as the sum of its connected components. 
\begin{definition}[Myerson value~\citep{myerson1977graphs} ] \label{def:m}
    Given a function $f$ and a graph $(V, E)$, the Myerson value of a node $i \in V$ is defined as
    \[
        \My_i =  \frac{1}{|V|} \sum_{T \subseteq V \setminus i} \frac{1}{{|V| - 1 \choose |T|}} \color{blue}{\left( f|_E(T \cup i) - f|_E(T)\right)}.
    \]
    \vspace{-3mm}
\end{definition}
The Myerson value is defined directly upon the Shapley value of the interaction-restricted function $f|_E$. Considering $f|_E$ instead of $f$ will only allow the interaction among connected nodes; thus, the Myerson values explicitly capture the graph structure information into the score $\psi_i$. By this definition, the Myerson value retains all the characteristics of the Shapley value. If $(V, E)$ is a complete graph, the Myerson value coincides with the Shapley value.
\section{Graph Explainer with Multiple Motifs}\label{sec:myerson}

The explanation task focuses on finding a subgraph $S \subseteq V$ so that the output of $f$ on $S$ is `most similar' to that of $f$ on the original input $V$. Cooperative game-based approaches to find $S$ often have two components: \textit{(i)} an allocation rule that attributes the importance scores to nodes or subsets of nodes in $V$, \textit{(ii)} an optimization model that takes the attribution scores and finds the optimal explanatory structures. 

For component (i), we introduce the Myerson-Taylor interaction index to capture structure information and high-order interactions in graph input. For component (ii), we propose an optimization model to identify the subgraph $S$. Figure~\ref{fig:sage} illustrates the overall flow of our method.

\subsection{Myerson-Taylor Interaction Index} \label{sec:mti}

%
    

We first propose the Myerson-Taylor index, which generalizes the Shapley value in both directions: capturing interactions and capturing the graph structure of the input.



\begin{figure*}
    \centering
    \begin{subfigure}[t]{0.52\textwidth}
        \centering        \includegraphics[width=\linewidth]{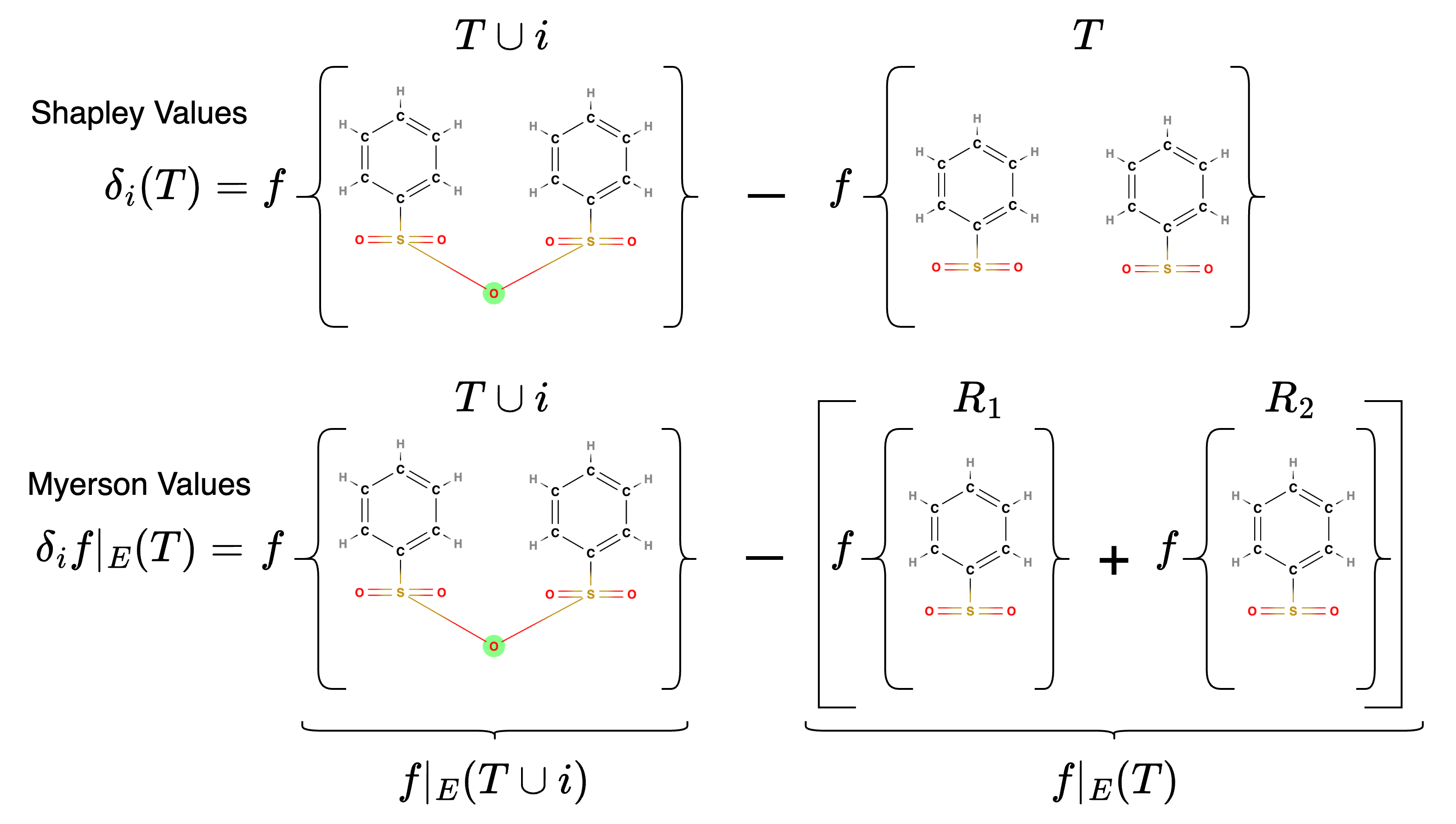}
        \vspace{-5mm}
        \caption{}
        \label{fig:shap}
    \end{subfigure}
    \begin{subfigure}[t]{0.43\textwidth}
        \includegraphics[width=\linewidth]{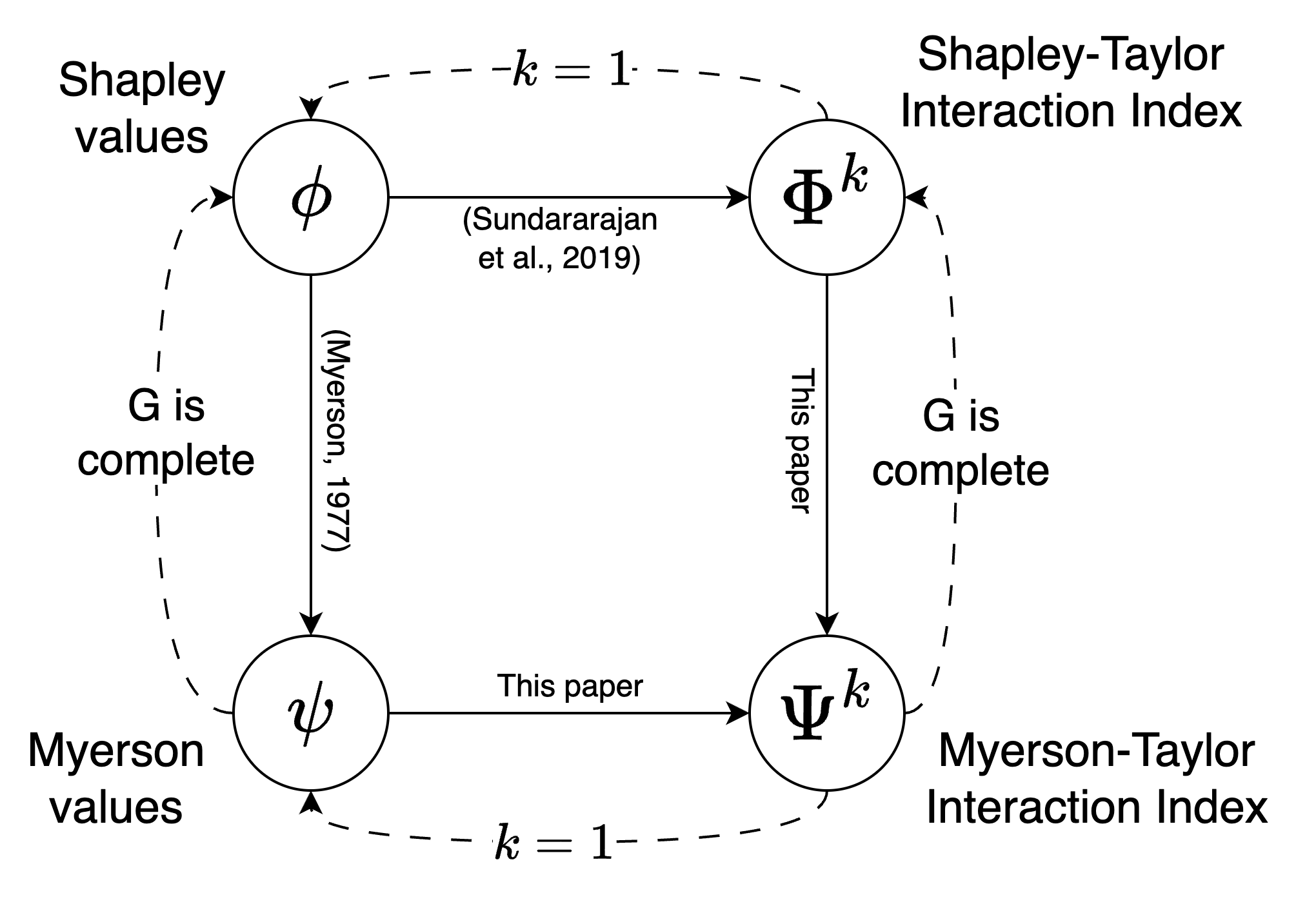}
        \vspace{-5mm}
        \caption{}
        \label{fig:myt_comp}
    \end{subfigure}
    \caption{(a) Examples of how Shapley and Myerson values evaluate a disconnected coalition. The set $T$ is not connected, and in the Myerson value, the function $f|_{E}(T)$ becomes the sum of output over two connected sets $R_1$ and $R_2$.  (b) The relations between the four allocation methods in this paper: solid arrows indicate the generalization direction, and dashed arrows indicate the recovery direction. Conditions for recovery are written on the dashed arrows.}
\end{figure*}

\begin{definition}[Myerson-Taylor index]\label{def:mti}
    Given a function $f$ and a graph $(V, E)$, the $k$-order Myerson-Taylor index of a subset $S \subseteq V$, $|S| \le k$ is defined as
    \[
        \MyI^k_S = \begin{dcases}
            \delta_S \blue{f|_E(\emptyset)} & \text{if}~|S| < k, \\
            \frac{k}{|V|} \sum_{T \subseteq V \setminus S} \frac{1}{{|V|-1 \choose |T|}} \delta_S \blue{f|_E(T)} & \text{if}~ |S| = k.
        \end{dcases}
    \]
\end{definition}
One can contrast Definition~\ref{def:mti} and~\ref{def:sht} to see that we replace the original model $f$ by the interaction-restricted function $f|_E$ that explicitly takes the graph structure of $(V, E)$ into consideration. Figure~\ref{fig:shap} shows how an interaction-restricted function $f|_E$ differs from the original function $f$ when evaluating a disconnected subgraph. This interaction-restricted function $f|_E$ is similar to the message-passing paradigm in GNNs in the sense that both only allow information propagation and aggregation among connected nodes. Thus, the role of $f|_E$ is to prevent the model from evaluating disconnected subgraphs, which could be pathological or OOD samples for the GNN models. In contrast, the Shapley-Taylor index $\Psi^k$ is structure-agnostic. 

In general, the Myerson-Taylor index generalizes from both the Myerson value and the Shapley-Taylor index to capture high-order interactions and structural information in the graph input. For a complete graph $(V, E)$, the Myerson-Taylor index recovers the Shapley-Taylor index, and for $k = 1$, it recovers the Myerson value (Figure~\ref{fig:myt_comp}).

\subsection{Motif Search}\label{sec:motif}
This section delineates the procedure to find multiple motifs that significantly sway the model’s predictions using the Myerson-Taylor interactions. It is crucial to note that our investigation extends beyond merely isolating a single motif that bolsters the model’s confidence score as in prior graph explainers~\citep{yuan2021explainability, zhang2022gstarx, ye2023same}. We also aim to uncover structures that obscure the model’s understanding, hindering it from giving a higher confidence score.


We define the space of possible explanations for the input graph $G$ as follows
\be \label{eq:H}
    \mc H_{m, M}\!=\!\left\{ \begin{array}{l}
    (S_1, \ldots, S_m) \subseteq V^m \text{ such that}: \\
    S_l \cap S_h = \emptyset ~\forall l,h,~
    \left|\cup_{l=1}^m S_l \right| \leq M \\
    S_l \text{ induces a \textit{connected} subgraph}
    \end{array}
    \right\}.
\ee
An $(m, M)$-explanation for the input $G$ is a decomposition of $G$ into $m$ subgraphs that are non-overlapping, of totally at most $M$ nodes, and each subgraph is a connected subgraph. Note that we do not impose a minimum node count on each motif, allowing for the possibility that $S_l$ could be empty and the number of highlighted motifs to fall below $m$. This obviates the need for users to explicitly calibrate $m$ in order to select an appropriate explanation. Therefore, both $m$ and $M$ serve as complexity budgets for the explanation. A higher $m$ allows for more dispersed explanations, and a higher $M$ enables explanations that include more nodes.




    
Let $\ve B \in \R^{n \times n}$ be a matrix capturing the second-order Myerson-Taylor ($\Psi^{2}$) between each node, \ie, $\ve B_{ij} = \MyI^2_{ij}$. Decompose $\ve B = \ve B^+ +\ve B^-$, where $\ve B^+ = \max(0, \ve B)$ and $\ve B^- = \min(0, \ve B)$ are matrices containing only positive and negative interactions, respectively. We define the Myerson-Taylor group attribution of a set $S$ as
\[
\mathrm{GrAttr}(S) = \sum_{\substack{i, j \in S \\ i \leq j}} \tau \ve B_{ij}^+ + (1 - \tau)\ve B_{ij}^-,
\]
where we explicitly constrain $i \le j$ to avoid double counting. The parameter $\tau \in [0, 1]$ allows users to focus on motifs that exert positive or negative contributions or both. 

We propose to extract the motifs from the solution of
\be \label{eq:reform}
    \max_{(S_1, \ldots, S_m) \in \mc H_{m, M}}~\displaystyle \sum_{l = 1}^m \Big| \mathrm{GrAttr}(S_l) \Big|.\
\ee

Problem~\eqref{eq:reform} maximizes the sum of \textit{absolute} group attribution values of identified motifs. The absolute operator ensures that negative interactions are also considered in the maximization. Ideally, nodes in the same motifs should strongly interact with each other either positively or negatively, while interactions of nodes from different motifs should be negligible. 

Problem~\eqref{eq:reform} is a variant of the quadratic multiple knapsack problem~\citep{hiley2006quadratic} with absolute values. One strategy to solve~\eqref{eq:reform} is by linear relaxations and then using off-the-shelf MILP solvers such as MOSEK~\citep{mosek} or GUROBI~\citep{gurobi}. The detailed discussions are provided in Appendix~\ref{sec:implementation}.

\subsection{Complexity Analysis}

The Myerson-Taylor index is easier to compute than the Shapley-Taylor index. The Shapley-Taylor index needs to evaluate $f$ for all possible subgraphs of $V$; however, the Myerson-Taylor index needs to evaluate $f$ only for all possible \textit{connected} subgraphs of $V$. While the number of connected subgraphs is still exponential in $|V|$, the number of queries can be significantly reduced for sparse graphs. This is an advantage of the Myerson-Taylor index when explaining large, sparse inputs or deep architectures. As a trade-off, the Myerson-Taylor index requires computing the connected components for evaluated subsets, which can be done in $\mc O(|V|)$ by the standard depth-first search algorithm. Similar to other game-based explainers~\citep{lundberg2017unified, yuan2021explainability, ye2023same, sundararajan2020shapley}, we also use Monte Carlo sampling to approximate the Myerson-Taylor index. 

Finally, MAGE is more computationally tractable compared to other cooperative-based graph explainers because it decomposes the attribution computation and subgraph search into two distinct phases. Thus, we only need to compute the interaction matrix $\ve B$ once for each input instance, and this can be done in parallel. In contrast, methods based on Monte Carlo Tree Search, like SubgraphX~\citep{yuan2021explainability} and SAME~\citep{ye2023same}, encounter a bottleneck due to the need for recalculating attribution scores for each motif candidate that is explored.


\section{Axiomatic Justification} \label{sec:axiom}

The Shapley value is theoretically attractive because it is unique under a specific set of axioms, ensuring a consistent scoring allocation as we change the model and the input. This property is thus desirable for its extensions~\cite{sundararajan2020shapley, myerson1977graphs, grabisch1999axiomatic}. We now provide a theoretical justification underpinning the Myerson-Taylor interaction index introduced in Section~\ref{sec:mti}. Let us first introduce a system of five axioms, which are inspired by the axioms that support the Shapley-Taylor interaction index~\citep{sundararajan2020shapley} and the Myerson value~\citep{selccuk2014axiomatization}. We recite the axioms for the Shapley-Taylor index in Appendix~\ref{sec:axiom_sht}.


\begin{axiom}[Linearity - \textbf{L}]
    A $k$-order interaction index $\Ic^k$ is linear, \ie, for any models $f_1, f_2$, a graph $G$, and a constant $\alpha$, we have $\Ic^k(f_1 + \alpha f_2, G) = \Ic^k(f_1, G) + \alpha \Ic^k(f_2, G)$, where $(f_1 + \alpha f_2)(T) = f_1 (T) + \alpha f_2 (T), \forall T \subseteq V$.
\end{axiom}


Linearity is a widely accepted axiom in the solution concepts of cooperative games, imposing additive behaviors to the allocation rules. We now define null nodes: a node $i \in V$ is a restricted null player if this node does not contribute to any coalitions it joins to form a \textit{connected} subgraph, \ie, $f(T \cup i) = \sum_{R \in \zeta(T)} f(R) + f(i)$ for any connected subset $T \cup i \subseteq V$.
\begin{axiom}[Restricted Null Player - {\bf RNP}]
    For a model $f$ and a graph $G = (V, E)$, let node $i \in V$ be a restricted null player, then the $k$-order interaction index $\mc I^k(f, G)$ satisfies
    \begin{enumerate}[leftmargin=5mm,label=(\roman*)]
        \item $\Ic^k_i (f, G) = f(i)$,
        \item for any $S \subseteq V, |S \cup i| \leq k$, we have $\Ic^k_{S \cup i} (f, G) = 0$.
    \end{enumerate}
\end{axiom}
This axiom resembles the \textit{dummy axiom} in the Shapley-Taylor interaction. However, instead of considering every possible subset of $V$, \bRNP~only focuses on \textit{connected} subgraphs of $V$ dictated by the edge information $E$. 
The axiom also implies that isolated nodes are inherently categorized as null players, suggesting they should not integrate with others to form motifs larger than their singular selves. The following axiom replaces \textit{the symmetry axiom} in the Shapley-Taylor index.

\begin{axiom}[Coalitional Fairness - {\bf CF}]
    A $k$-order interaction index $\Ic^k$ is coalitional fair  for a graph $G$ if for any connected coalition $T$, \ie, $|\zeta(T)| = 1$, and two models $f_1$ and $f_2$ such that $f_1(R) = f_2(R)$ for all $R \neq T$, we then have
    \begin{align*}
        \Ic^k_{S_1}(f_1, G) - \Ic^k_{S_1}(f_2, G) = \Ic^k_{S_2}(f_1, G) - \Ic^k_{S_2}(f_2, G),
    \end{align*}
    for any $S_1, S_2 \subseteq T$ such that $|S_1| = |S_2|$.
\end{axiom}
Coalitional fairness dictates that a change in the value of a connected coalition should result in an equitable redistribution of interaction levels across all subsets of equivalent size within that coalition~\citep{selccuk2014axiomatization}.

The next axiom requires the definition of unanimity functions. A function $u_T$ for a $T \in V$ is \textit{unanimity} function if the formation of the coalition $T$ is necessary and sufficient for $u_T$ to have non-zero value:
\[
    u_T(S) = \begin{dcases}
        1 & \text{if}~ S \supseteq T, \\
        0 & \text{otherwise}.
    \end{dcases}
\]

\begin{axiom}[Interaction Distribution - {\bf ID}]
    A $k$-order interaction index $\Ic^k$ satisfies \bID~ if for any unanimity function $u_T$, and a graph $(V, E)$ in which $T \subseteq V$ is connected, we have
    \[
        \Ic^k_S (u_T, G) = 0,
    \]
    for all $S \subsetneq T$ such that $|S| < k$.
\end{axiom}
Similar to~\citep{sundararajan2020shapley}, the axiom \bID~ensures the lower interaction orders ($l < k$) cannot be captured by $k$-th order interactions and vice versa. Meanwhile, $k$-th order interaction of a set $S$ with $|S| = k$ will capture interactions of $S$ and its supersets ($\forall T \supsetneq S$). \bID~is used to introduce the efficiency axiom for the Shapley-Taylor interaction index, arguably the main advantage of the Shapley-Taylor index compared to the classical Shapley interaction index~\citep{grabisch1999axiomatic}. We also expect a similar axiom for the Myerson-Taylor index. 

\begin{axiom}[Component Efficiency - \bCE]
    A $k$-order interaction index $\Ic^k$ is component efficient if, for any graph $G=(V, E)$ and any model $f$, we have
    \[
        \sum_{S \subseteq C, |S| \leq k} \Ic^k_{S}(f, G) = f(C) - f(\emptyset) \quad \forall C \in \zeta (V).
    \]
\end{axiom}
The \bCE~axiom ensures that the confidence score of the model is fully and fairly distributed among its interacting components. In case the graph $G$ is connected, \bCE~coincides with the efficiency axiom of the Shapley-Taylor interaction, \ie, $\sum_{S\subseteq V, |S| \leq k} \mc I_S^k(f, G) = f(V) - f(\emptyset)$.

To justify the Myerson-Taylor index, we show that it is a unique construction that can satisfy the above five axioms. 

\begin{theorem}[Uniqueness]
    \label{thm:mt_uni}
    The Myerson-Taylor index is the unique interaction allocation rule that satisfies \bL, \bRNP, \bCF, \bID, and \bCE~axioms.
\end{theorem}
This result emphasizes the importance of the Myerson-Taylor index, as it uniquely extends the Myerson value and Shapley-Taylor index to adhere to the five outlined axioms that account for structural information and high-order node interactions. The proof is relegated to Appendix~\ref{sec:proofs}.

It is worth noting that the notion of coalition fairness in our axiom system aligns with the four-axiom system of the Myerson value proposed in~\citep{selccuk2014axiomatization} instead of the fairness notion in the original work~\citep{myerson1977graphs}. In Appendix~\ref{sec:additional_results}, we generalize the classical fairness axiom~\citep{myerson1977graphs} to higher-order interactions and show that the Myerson-Taylor index also complies with this extended fairness criterion.

\begin{figure*}
\begin{subfigure}[t]{0.485\textwidth}
    \centering
    \includegraphics[width=\linewidth]{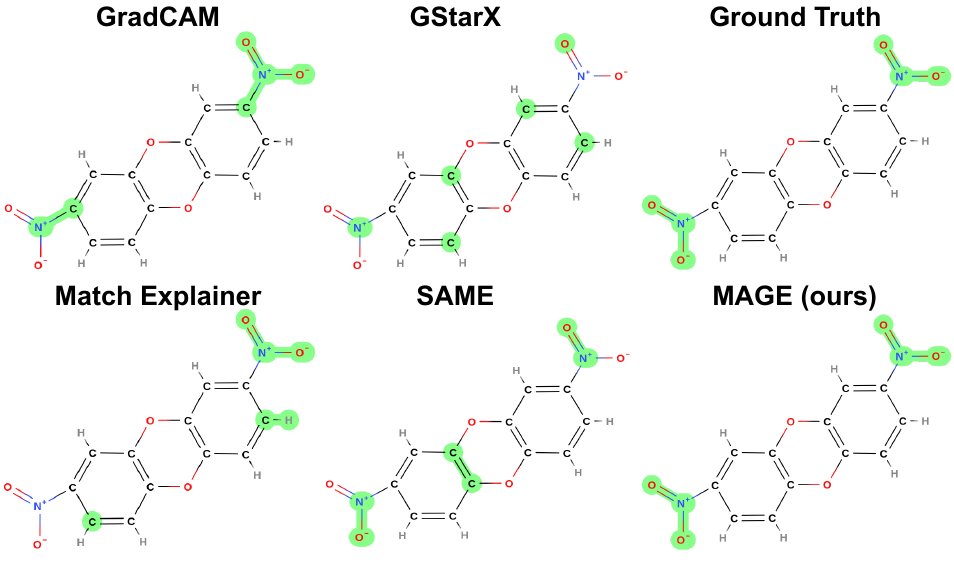}
    \caption{}
    \label{fig:multi_motif_ex}
\end{subfigure}
\hfill
\begin{subfigure}[t]{0.315\textwidth}
    \centering
    \includegraphics[width=\linewidth]{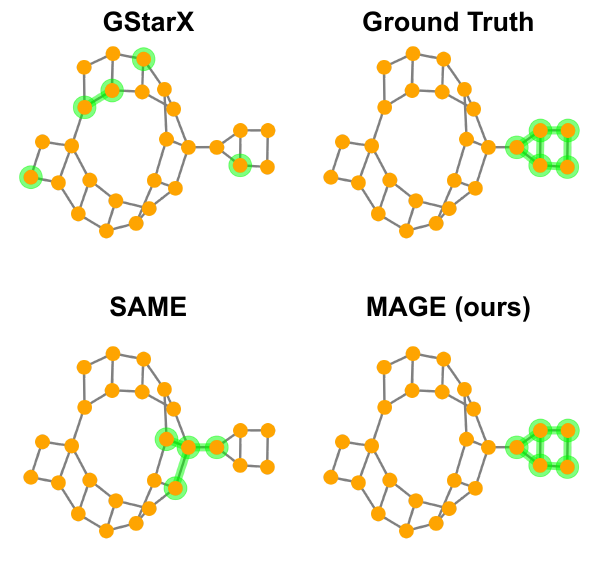}
    \caption{}
    \label{fig:sigle_motif_ex}
\end{subfigure}
\hfill 
\begin{subfigure}[t]{0.166\textwidth}
    \centering
    \includegraphics[width=\linewidth]{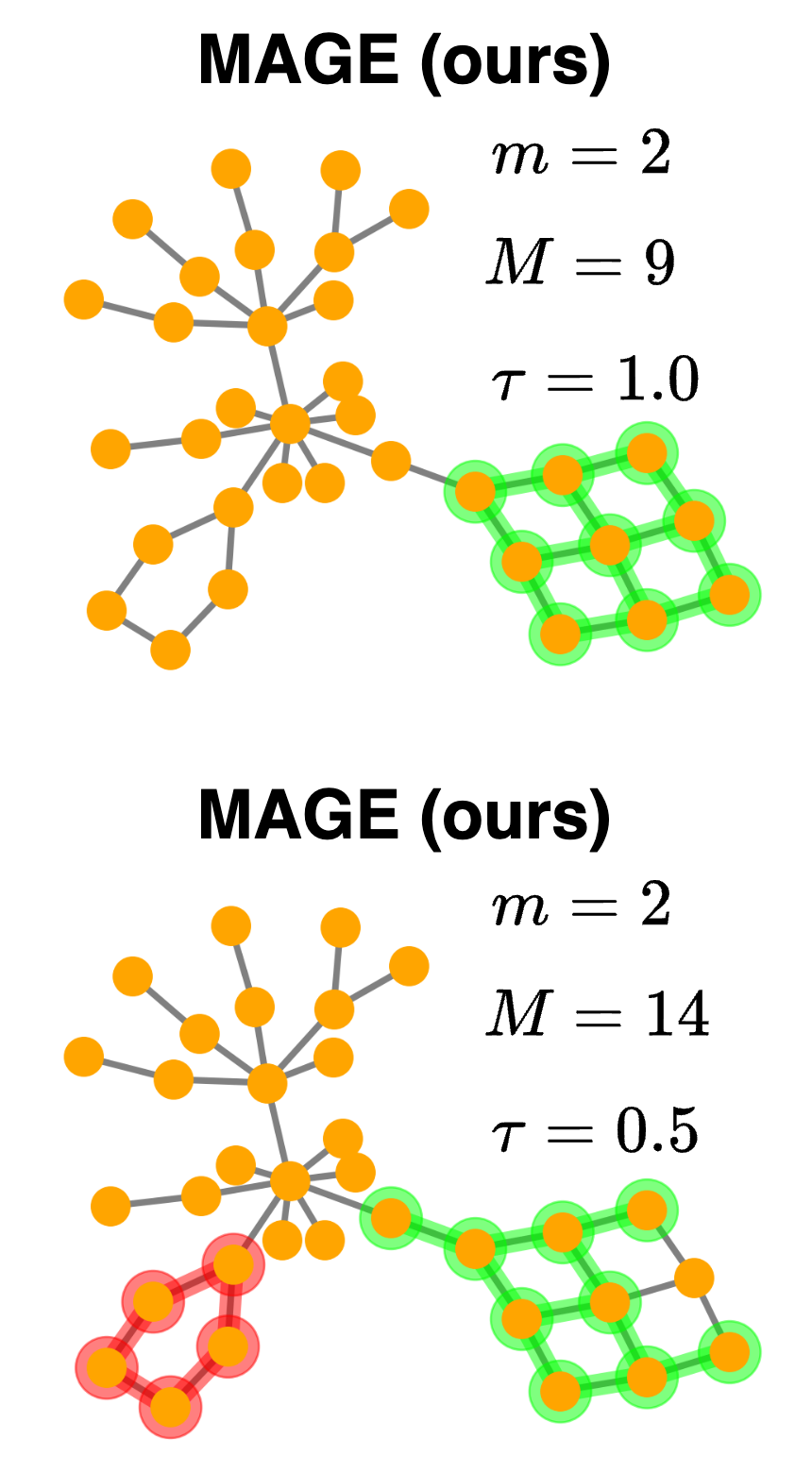}
    \caption{}
    \label{fig:neg_motif}
\end{subfigure}
\vspace{-2mm}
\caption{(a) An example in the Mutagenic dataset. Only MAGE correctly highlights the two -NO2 groups. (b) An example in the SPMotif dataset. Only MAGE can identify the house motif in the input graph. (c) An example in BA-HouseGrid shows MAGE's ability to highlight negative motifs. Green indicates positive motifs, and red indicates negative motifs. (model prediction: grid)}
\end{figure*}

\begin{table*}[ht]
    \caption{Results for single motif GCN \& multiple motifs GIN. On average, MAGE achieves a 59.29\% improvement in F1 on single motif datasets, a 28.11\% improvement in AMI on multi-motif datasets, and a 12.61\% improvement in AUC across all datasets.} 
    \label{tab:with_ground_truth}
    \resizebox{\textwidth}{!}{
    \begin{filecontents*}{single_gcn_multi_gin_no_std_improv.csv}
methods,BA-2Motifs,BA-2Motifs,BA-HouseGrid,BA-HoseGrid,SP-Motif,SP-Motif,MNIST-75SP,MNIST-75SP,BA-HouseAndGrid,BA-HouseAndGrid,BA-HouseOrGrid,BA-HouseOrGrid,Mutagenic,Mutagenic,Benzene,Benzene
,\textit{F1}$\uparrow$,\textit{AUC}$\uparrow$,\textit{F1}$\uparrow$,\textit{AUC}$\uparrow$,\textit{F1}$\uparrow$,\textit{AUC}$\uparrow$,\textit{F1}$\uparrow$,\textit{AUC}$\uparrow$,\textit{AMI}$\uparrow$,\textit{AUC}$\uparrow$,\textit{AMI}$\uparrow$,\textit{AUC}$\uparrow$,\textit{AMI}$\uparrow$,\textit{AUC}$\uparrow$,\textit{AMI}$\uparrow$,\textit{AUC}$\uparrow$
GradCAM,\cellcolor{yellow!25}\underline{0.634},\cellcolor{yellow!25}\underline{0.753},0.459,0.485,\cellcolor{yellow!25}\underline{0.491},\cellcolor{yellow!25}\underline{0.616},0.193,0.492,\cellcolor{yellow!25}\underline{0.825},\cellcolor{green!25}\textbf{0.994},\cellcolor{yellow!25}\underline{0.931},\cellcolor{green!25}\textbf{0.997},-0.001,0.514,\cellcolor{yellow!25}\underline{0.789},\cellcolor{green!25}\textbf{0.964}
GNNExplainer,0.222,0.440,0.297,0.546,0.185,0.465,0.220,0.531,0.275,0.533,0.148,0.532,0.228,0.679,0.178,0.487
PGExplainer,0.042,0.498,0.057,0.434,0.066,0.097,0.236,\cellcolor{yellow!25}\underline{0.607},0.100,0.088,0.170,0.002,0.099,0.573,0.186,0.042
Refine,0.144,0.474,0.191,0.398,0.164,0.508,0.153,0.459,0.254,0.429,0.123,0.488,0.210,0.623,0.207,0.529
MatchExplainer,0.586,0.706,\cellcolor{yellow!25}\underline{0.587},\cellcolor{yellow!25}\underline{0.712},0.190,0.513,0.162,0.483,0.537,\cellcolor{yellow!25}\underline{0.810},0.521,\cellcolor{yellow!25}\underline{0.788},0.216,0.576,0.318,0.545
SubgraphX,0.620,0.720,0.480,0.700,\cellcolor{green!25}\textbf{0.542},\cellcolor{yellow!25}\underline{0.680},0.170,0.501,0.494,0.697,0.526,0.767,\cellcolor{yellow!25}\underline{0.595},\cellcolor{yellow!25}\underline{0.784},0.731,\cellcolor{yellow!25}\underline{0.832}
GStarX,0.180,0.480,0.267,0.544,0.203,0.498,\cellcolor{yellow!25}\underline{0.280},0.517,0.203,0.494,0.130,0.484,-0.018,0.462,0.122,0.505
SAME,0.630,0.730,0.474,0.693,0.410,0.610,\cellcolor{yellow!25}\underline{0.272},0.531,0.497,0.681,0.606,\cellcolor{yellow!25}\underline{0.796},0.480,0.709,0.617,0.791
MAGE (ours),\cellcolor{green!25}\textbf{0.858},\cellcolor{green!25}\textbf{0.890},\cellcolor{green!25}\textbf{0.832},\cellcolor{green!25}\textbf{0.849},\cellcolor{green!25}\textbf{0.547},\cellcolor{green!25}\textbf{0.699},\cellcolor{green!25}\textbf{0.634},\cellcolor{green!25}\textbf{0.716},\cellcolor{green!25}\textbf{0.998},\cellcolor{green!25}\textbf{0.999},\cellcolor{green!25}\textbf{0.998},\cellcolor{green!25}\textbf{0.999},\cellcolor{green!25}\textbf{1.000},\cellcolor{green!25}\textbf{1.000},\cellcolor{green!25}\textbf{0.917},\cellcolor{green!25}\textbf{0.959}
\textbf{Improvement (\%)},35.33,18.19,41.74,19.24,0.92,2.79,126.43,17.96,20.97,0.50,7.20,0.20,68.07,27.55,16.22,-0.52
\end{filecontents*}
\pgfplotstabletypeset[
        col sep=comma,
        string type,
        every head row/.style={before row=\toprule},
        every row no 0/.style={after row=\midrule},
        every head row/.style={output empty row, before row={%
                \toprule \multirow{3}{*}{Method} &
                \multicolumn{8}{c}{Single Motif - GCN} & 
                \multicolumn{8}{c}{Multiple Motifs - GIN}\\ 
                \cmidrule(r){2-9} \cmidrule(r){10-17}
                &\multicolumn{2}{c}{BA-2Motifs} & 
                \multicolumn{2}{c}{BA-HouseGrid} & 
                \multicolumn{2}{c}{SPMotif} & 
                \multicolumn{2}{c}{MNIST75SP} & 
                \multicolumn{2}{c}{BA-HouseAndGrid} & \multicolumn{2}{c}{BA-HouseOrGrid} & \multicolumn{2}{c}{Mutagenic} & \multicolumn{2}{c}{Benzene} \\
                \cmidrule(r){2-3} \cmidrule(r){4-5} \cmidrule(r){6-7} \cmidrule(r){8-9} \cmidrule(r){10-11} \cmidrule(r){12-13} \cmidrule(r){14-15} \cmidrule(r){16-17}
            }},
        every row no 1/.style={after row=\tabucline[0.4pt blue!40 off 2pt]{-}},
        every row no 5/.style={after row=\tabucline[0.4pt blue!40 off 2pt]{-}},
        every last row/.style={before row=\midrule, after row=\bottomrule},      
    ]{single_gcn_multi_gin_no_std_improv.csv}
    }
\end{table*}

\begin{table}[t]
\centering
\caption{Fidelity evaluation on sentiment classification and GCN. $\#Q$ denotes the number of GNN queries needed.}
\label{tab:sst_gcn}
\scriptsize
\begin{filecontents*}{sst_gcn_no_std_queries.csv}
methods,GraphSST2,GraphSST2,GraphSST2,Twitter,Twitter,Twitter
,$\mathrm{Fid}_{\alpha}$$\uparrow$,$\mathrm{Fid}$$\uparrow$,$\mathrm{\#Q}$$\downarrow$,$\mathrm{Fid}_{\alpha}$$\uparrow$,$\mathrm{Fid}$$\uparrow$,$\mathrm{\#Q}$$\downarrow$
GradCAM,\cellcolor{yellow!25}\underline{0.169},0.253,N/A,0.268,0.363,N/A
SubgraphX,0.141,0.225,255K,0.238,0.32,312K
GStarX,\cellcolor{yellow!25}\underline{0.161},\cellcolor{yellow!25}\underline{0.273},\cellcolor{yellow!25}\underline{17K},0.276,\cellcolor{yellow!25}\underline{0.425},\cellcolor{yellow!25}\underline{20K}
SAME,0.141,0.216,24K,\cellcolor{yellow!25}\underline{0.293},0.397,31K
MAGE (ours),\cellcolor{green!25}\textbf{0.200},\cellcolor{green!25}\textbf{0.337},\cellcolor{green!25}\textbf{2K},\cellcolor{green!25}\textbf{0.317},\cellcolor{green!25}\textbf{0.471},\cellcolor{green!25}\textbf{3K}
\end{filecontents*}
\pgfplotstabletypeset[
    col sep=comma,
    string type,
    every head row/.style={before row=\toprule},
    every row no 0/.style={after row=\midrule},
    every head row/.style={output empty row, before row={%
            \toprule \multirow{2}{*}{Method}  &
            \multicolumn{3}{c}{GraphSST2} & \multicolumn{3}{c}{Twitter} \\
            \cmidrule(r){2-4} \cmidrule(r){5-7}
        }},
    every row no 1/.style={after row=\tabucline[0.4pt blue!40 off 2pt]{-}},
    every last row/.style={after row=\bottomrule},
]{sst_gcn_no_std_queries.csv}
\end{table}

\section{Experiments}

We evaluate our method, Myerson-Taylor Structure-Aware Graph Explainer (MAGE)\footnote{Our implementation is available at: \url{https://github.com/ngocbh/MAGE/}}, on ten datasets and three GNN models and compare it with eight baselines to show the effectiveness of MAGE in identifying explanatory structures. 

\noindent\textbf{Datasets.} We use ten datasets commonly used in the graph explainability literature, including synthetic data, biological, text, and image data. For \textit{synthetic datasets}, we use Ba-2Motifs~\citep{luo2020parameterized}, BA-HouseGrid~\citep{amara2023ginx}, and SPMotif~\citep{wu2022discovering} for classification tasks involving Barabási base structures with distinct motifs. 
For \textit{molecular property prediction}, we use Mutagenic~\citep{kazius2005derivation} and \textit{Benzene}~\citep{sanchez2020evaluating}. Molecular graphs are labeled based on their property, and the chemical fragments (-NO2 and -NH2 for Mutagenic and benzene rings for Benzene) are identified as ground-truth explanations. For \textit{image classification}, we use MNIST75SP~\citep{monti2017geometric}, where each image in MNIST is transformed into a graph of superpixels, with edges defined by the spatial neighborhood of the superpixels. And for \textit{sentiment classification}, we employ two datasets \textit{GraphSST2} and \textit{Twitter}~\citep{yuan2022explainability} where each node corresponds to one word in the text, edges are constructed by the Biaffine parser~\citep{gardner2018allennlp}.

Notably, BA-2Motifs, BA-HouseGrid, SPMotif, and MNIST75SP have only one explanatory structure within a graph, while graphs in BA-HouseAndGrid, BA-HouseOrGrid, Mutag, and Benzene may have multiple explanatory structures. GraphSST2 and Twitter do not have ground truth explanations. Full descriptions of datasets are provided in Appendix~\ref{sec:app_data}.

\noindent\textbf{Models.} We use three popular GNNs: GCN~\citep{kipf2016semi}, GIN~\citep{xu2018powerful}, and GAT~\citep{velivckovic2017graph}. We report the accuracy and hyperparameters in the Appendix~\ref{sec:app_model}. As GAT performs poorly on synthetic data, we only explain for GAT on real-world data. 



\noindent\textbf{Baselines.} We use seven common baselines in perturbation-based graph explainability, including \textit{GNNExplainer}~\citep{ying2019gnnexplainer}, \textit{PGExplainer}~\citep{luo2020parameterized}, \textit{Refine}~\citep{wang2021towards}, \textit{MatchExplainer}~\citep{wu2023explaining}, \textit{SubgraphX}~\citep{yuan2021explainability}, \textit{GStarX}~\citep{zhang2022gstarx}, \textit{SAME}~\citep{ye2023same}. SubgraphX, GStarX, and SAME are cooperative game-based explainers; thus, they are in the same category as our method. Moreover, we also compare MAGE against \textit{GradCAM}~\citep{pope2019explainability}, a white-box gradient-based explainer adapted to explain GNNs. 

\noindent\textbf{Metrics.}  We use the standard metrics for explanation tasks:
\begin{itemize}[topsep=-4px,partopsep=-4px, parsep=-4px]
    \item For datasets with ground truth explanations, we evaluate the accuracy of explanations using the \textit{F1 score}, \textit{Adjusted Mutual Information (AMI)}, and \textit{Area Under the Curve (AUC)}. The F1 score reports the overlap of the nodes highlighted by the explainers compared to the ground truth. For datasets with multiple motifs, we use the AMI score, a widely used metric in clustering tasks, to measure the explainer's ability to identify different motifs in the graph structure. Following practice in ~\citep{ying2019gnnexplainer, luo2020parameterized}, we also report the AUC score by comparing edge masks generated by explainers against the ground truth edge masks.
    \item For datasets with\textit{out} ground-truth explanations, we utilize the \textit{Fidelity} ($\mathrm{Fid}$)~\citep{yuan2022explainability} to measure the faithfulness of explanations to the model's prediction. Because $\mathrm{Fid}$ is sensitive to OOD samples, thus favoring OOD explanations~\citep{zheng2023towards, amara2023ginx}, we also measure $\mathrm{Fid_{\alpha}}$ proposed in~\citep{zheng2023towards} to alleviate the OOD problem of $\mathrm{Fid}$. 
\end{itemize}
Appendix~\ref{sec:app_metric} provides details for the above metrics.


\begin{figure*}[!ht]
    \centering    \includegraphics[width=\linewidth]{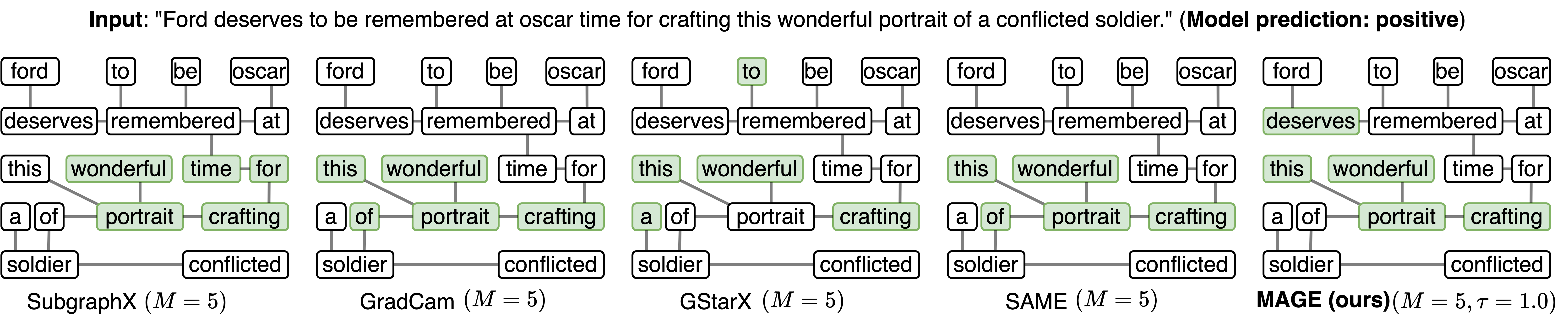}
    \caption{An example in the Graph-SST2 dataset. MAGE's explanation is more concise and correctly captures the main verb `deserves', crucial to determining a sentence's sentiment, while other baselines fail to identify it.} 
    \label{fig:sst_ex}
\end{figure*}
\noindent\textbf{Setup and implementation details.} We split the dataset into training, validation, and test subsets with respective ratios of 0.8, 0.1, and 0.1. We train GNN models to a reasonable performance and then run the explainers for graph instances in the test datasets. We report the average metrics over instances in the test dataset.

Regarding hyperparameter settings, we set the number of explanatory nodes $M$ and components $m$ according to the ground truth explanations for all the baselines if they are available. For the datasets without ground truth (sentiment classification), we set $M$ to be 30\% of the number of nodes in the graph. We set $\tau=1$ for our method as ground-truth explanations, and baselines are only for motifs with positive contributions. The number of permutations used to compute the Myerson-Taylor index is set to $200$, and we use MOSEK~\citep{mosek} with default parameters for the motif search.  All the results are averaged over five times tests with different random seeds.

\subsection{Quantitative Results} 

We report the results for datasets with ground truth explanations in Table~\ref{tab:with_ground_truth}. Our method demonstrates superior performance to all baselines by achieving a 12.51\% improvement in the AUC metric. MAGE improves the F1 score by 58.64\% for datasets with a single motif. For multi-motif datasets, MAGE also improves the AMI score by 28.11\% compared to the baselines, indicating more faithful explanations in datasets with multiple explanatory structures. Notably, MAGE accurately identifies all -NO2 and -NH2 chemical groups contributing to a molecule's mutagenic property for the GCN model. Compared to GradCAM, a white-box gradient-based method, MAGE's explanations are better than GradCAM by 17\% in the AUC and 65\% in the F1 score. For sentiment classification tasks without ground truth (Table~\ref{tab:sst_gcn}), MAGE achieves a 14\% higher fidelity score than the best baselines. Moreover, Table~\ref{tab:sst_gcn} also shows that MAGE is much more query-efficient than other game-based methods. Additional results for other models and running time analysis are provided in the appendix. 



\begin{table}[t]
\centering
\caption{Ablating Shapley-Taylor ($\Phi^2$) and Myerson-Taylor ($\Psi^2$) indices and connectivity constraints in the problem~\eqref{eq:reform}.}
\label{tab:ablation}
\scriptsize
\begin{filecontents*}{abla_cnvt_gcn.csv}
methods,BA-2Motifs,BA-2Motifs,BA-HouseGrid,BA-HouseGrid
,\textit{F1} $\uparrow$,\textit{AUC}$\uparrow$,\textit{F1} $\uparrow$,\textit{AUC}$\uparrow$
MAGE ($\Phi^2$) w/o connectivity,0.699,0.773,0.634,0.735
MAGE ($\Phi^2$) w/ connectivity,\cellcolor{yellow!25}\underline{0.709},\cellcolor{yellow!25}\underline{0.787},0.636,0.734
MAGE ($\Psi^2$) w/o connectivity,\cellcolor{green!25}\textbf{0.854},\cellcolor{green!25}\textbf{0.885},\cellcolor{yellow!25}\underline{0.819},\cellcolor{yellow!25}\underline{0.838}
MAGE ($\Psi^2$) w/ connectivity,\cellcolor{green!25}\textbf{0.858},\cellcolor{green!25}\textbf{0.890},\cellcolor{green!25}\textbf{0.832},\cellcolor{green!25}\textbf{0.849}
\end{filecontents*}
\pgfplotstabletypeset[
    col sep=comma,
    string type,
    every head row/.style={before row=\toprule},
    every row no 0/.style={after row=\midrule},
    every head row/.style={output empty row, before row={%
            \toprule \multirow{2}{*}{Method}  &
            \multicolumn{2}{c}{BA-2Motifs} & \multicolumn{2}{c}{BA-HouseGrid} \\
            \cmidrule(r){2-3} \cmidrule(r){4-5}
        }},
    every last row/.style={after row=\bottomrule},
]{abla_cnvt_gcn.csv}
\vspace{-3mm}
\end{table}

\subsection{Qualitative Results} 

We show qualitative results for different tasks in Figure~\ref{fig:intro_example}, and Figures~\ref{fig:multi_motif_ex}-\ref{fig:neg_motif}. More examples are in Appendix~\ref{sec:app_result}. 

\noindent\textbf{Positive motifs.} For topology-based tasks, as shown in Figure~\ref{fig:sigle_motif_ex}, MAGE accurately identifies the house motif that represents the graph class. For molecular classification tasks, including Mutagenic and Benzene,  MAGE can highlight all chemical groups that exist in the molecular structure (e.g., -NO2 and -NH2 in Mutagenic shown in Figure~\ref{fig:multi_motif_ex}, and carbon rings in Benzene in Figure~\ref{fig:intro_example}). Thus, the structures highlighted by MAGE align with the ground truth explanations. In contrast, other baselines struggle to provide meaningful explanations in these cases. Moreover, Figure~\ref{fig:sst_ex} shows an example from sentiment classification where MAGE adeptly highlights the main verb `deserves' in a sentence, which is crucial for assessing the sentence's overall sentiment. Appendix~\ref{sec:app_qualitative} provides more visualizations for the remaining datasets.  

\noindent\textbf{Negative motifs.} Figure~\ref{fig:neg_motif} shows that MAGE can identify substructures with negative contributions to the prediction. We examine a GIN model trained on the BA-HouseGrid dataset to classify house and $3 \times 3$ grid motifs. We explain the model's prediction for a grid example with a manually injected five-cycle motif into the structure. When we set $\tau = 1.0$, MAGE can correctly identify the grid structure, which causes the model prediction. As we calibrate $\tau = 0.5$ to consider both negative and positive contributions, MAGE can effectively highlight five-cycle and grid motifs. Here, a five-cycle structure may mislead the model's prediction towards a label for the house motif.

\subsection{Ablation Study}

This experiment shows the usefulness of structural information in graph explainer. \textit{First}, we ablate the second-order Myerson-Taylor index in MAGE with the second-order Shapley-Taylor. \textit{Second}, we run MAGE without enforcing the connectivity constraint for explanatory structures ($S_l$) in the optimization problem~\eqref{eq:reform}. Other settings are the same as in Table~\ref{tab:with_ground_truth}. Table~\ref{tab:ablation} shows that the Myerson-Taylor index improves the MAGE's explanation significantly compared to the Shapley-Taylor, and the connectivity constraint also improves the explanation accuracy. Connectivity constraint for each motif is important to avoid fragmented explanations, which are hard to interpret for humans. 

We also conduct an ablation study to compare the group attribution approximation of four methods, Shapley, Myerson, second-order Shapley-Taylor, and second-order Myerson-Taylor on Imagenet with ResNet50 and ViT16 models. We leave this experiment to the appendix.

\section{Conclusion}

This paper introduced the Myerson-Taylor index, which captures high-order interactions and the graph structure to explain GNN models. We proposed MAGE, a graph explainer that leverages the second-order Myerson-Taylor index to compute the motifs' attributions and highlight ones that are influential to the GNN's prediction, both positively and negatively. Extensive experiments on various domains show the compelling results of MAGE compared to other baselines.

\textbf{Limitation and future work.} Our approach, similar to other game-based explainers, offers consistent, stable, and model-agnostic importance scores but at the expense of high computational costs. Exact computations for these methods require exponential time with respect to the number of nodes ($|V|$) in a graph. In practice, we usually need to approximate the importance scores using Monte Carlo sampling. Notably, MAGE outperforms other game-based explainers for GNNs in terms of efficiency, requiring fewer model queries and offering faster run times.

Further, our Myerson-Taylor index treats GNN models as black-box models, thus permitting unrestricted information sharing among nodes in a connected component. This feature may not align with the practical, layer-restricted information propagation in GNNs. Addressing this discrepancy remains an open question. Moreover, as the Myerson-Taylor index uniquely generalizes the Myerson and Shapley-Taylor indices to include network structures and retains all their characteristics, extensions of the Myerson-Taylor index to study the interactions of players in (weighted) network games would be a potential research direction.

\section*{Acknowledgement}
This project is made possible through the generous support of Snap Inc.

\section*{Impact Statement}
This paper primarily focuses on methods for explaining graph neural network (GNN) predictions, aiming to enhance the interpretability of black-box models. The datasets we utilize are
publicly available. Our work has many potential societal consequences, none of which must be specifically highlighted here.

\bibliography{main.bbl}
\bibliographystyle{icml2024}

\newpage
\appendix
\onecolumn
\section{Axioms of the Shapley-Taylor Interaction Index}\label{sec:axiom_sht}
For comparison purposes, we provide the axiomatic characterizations of the Shapley-Taylor interaction index~\citep{sundararajan2020shapley}.

\begin{axiom}[Linearity - \textbf{L}]
    \label{axm:shi-l}
    $\mc I^k$ is a linear function, \ie, for any two functions $f_1, f_2$, we have $\Ic^k(f_1 + \alpha f_2) = \Ic^k(f_1) + \alpha \Ic^k(f_2)$, for any constant $\alpha$.
\end{axiom}

\begin{axiom}[Dummy - \textbf{D}]
    \label{axm:shi-d}
    If $i$ is a dummy node, \ie, for any subset $T \subseteq V \setminus i$, $f(T \cup i) = f(T) + f(i)$, then
    \[
        \mc I^k_{S \cup i} = 
        \begin{dcases}
            f(i) & \text{if}~~S = \emptyset, \\
            0 & \text{if}~~S \subseteq V, |S| \leq k - 1.  \\
        \end{dcases}
    \]
\end{axiom}

\begin{axiom}[Symmetry - \textbf{S}]
    \label{axm:shi-s}
    For any permutation order $\pi$ on $V$, we have $\Ic^k_S(f) = \mc I_{\pi S}(\pi f)$ where $\pi f(T) = f(\pi T)$.
\end{axiom}

\begin{axiom}[Interaction Distribution - \textbf{ID}]
    \label{axm:shi-id}
    For a unanimity function $u_T$, for all $S \subsetneq T, |S| < k$, we have $\mc I^k_S(u_T) = 0$.
\end{axiom}

\begin{axiom}[Efficiency - \textbf{E}]    
\label{axm:shi-e}
For any function $f$, we have
    \[
        \sum_{S \subseteq V, |S| \leq k} \mc I^k_S (f) = f(V).
    \]
\end{axiom}

\begin{theorem}[Shapley-Taylor uniqueness~\citep{sundararajan2020shapley}]
    Shapley-Taylor index is the only interaction index that satisfies five axioms: \bL, \bD, \bS, \bE, and \bID.
\end{theorem}
While there exist several methods to compute interaction among players~\citep{grabisch1999axiomatic}, the key attraction of the Shapley-Taylor interaction index is its satisfaction with the efficiency axiom (Axiom~\ref{axm:shi-e}). The idea of the Shapley-Taylor index is drawn from the Taylor expansion of the multilinear extension of a cooperative game~\citep{owen1972multilinear, sundararajan2020shapley}. It posits that the interactions within a subset $S$ of size $l$ less than $k$ are analogous to the $l$-th order term in a Taylor series, capturing only the interactions inherent to that subset. On the other hand, the $k$-order Shapley-Taylor indices are akin to the Lagrange remainder of the series, encompassing both the interactions of the set itself and those of higher orders.

\section{Proofs}\label{sec:proofs}
This section provides the detailed proofs for Theorem~\ref{thm:mt_uni}.

Let $\mc F^V$ denote a space of functions acting on the set of vertices $V$ without restriction
\[
    \mc F^V = \{f: 2^V \to \R\},
\]
and $\mc F^V|_E$ be a space of interaction-restricted functions on the graph $(V, E)$
\[
    \mc F^V|_E = \left\{ f|_E = \sum_{R \in \zeta(T)} f(R) \text{~where~} f \in \mc F\right\}. 
\]
Thus, the space of interaction-restricted function $\mc F^V|_E$ is a subspace of $\mc F$.

For convenience, we present a definition of the Myerson-Taylor index using the Shapley-Taylor index as follows.

\begin{definition}[Myerson-Taylor index]\label{def:redef-mti}
    Given a function $f$ and a graph $(V, E)$, the $k$-order Myerson-Taylor index of a subset $S \subseteq V$, $|S| \le k$ is
    \[
        \MyI^k_S(f, E) = \ShI^k_S (f|_E),
    \]
    where $\ShI^k_S$ is the Shapley-Taylor index of $S$ with respect to the interaction-restricted function $f|_E$.
\end{definition}
By Definition~\ref{def:redef-mti}, the Myerson-Taylor index is defined directly upon the Shapley-Taylor index, thus inherently retaining all characteristics of the Shapley-Taylor index for interaction-restricted functions.

To prove Theorem~\ref{thm:mt_uni}, we first need to show the Myerson-Taylor index satisfies five axioms, which can be done by combining definitions of the Myerson-Taylor index and interaction-restricted functions.

\begin{proposition} \label{prop:myti-l}
    The Myerson-Taylor index satisfies the linearity (\bL) axiom.
\end{proposition}
    
\begin{proof}[Proof of Proposition~\ref{prop:myti-l}] 
For a graph $(V, E)$ and two functions $f_1, f_2 \in \mc F^V$, we have
\begin{align*}
    (f_1 + \alpha f_2)|_E (T) &= \sum_{R \in \zeta(T)} (f_1 + \alpha f_2)(R) \\ 
    &= \sum_{R \in \zeta(T)} f_1(R) + \sum_{R \in \zeta(T)} \alpha f_2(R) \\
    &= f_1|_E(T) + \alpha f_2|_E (T).
\end{align*}
Thus, the communication restriction preserves the linearity. 

As the Myerson-Taylor index is defined upon the Shapley-Taylor index, which is also a linear function on $\mc F^V$, the Myerson-Taylor index is a linear function on $\mc F^V$.
\end{proof}

\begin{proposition} \label{prop:myti-rnp}
    The Myerson-Taylor index satisfies the restricted null player (\bRNP) axiom.
\end{proposition}
    
\begin{proof}[Proof of Proposition~\ref{prop:myti-rnp}] 
We first show that for any restricted null node $i$ of the graph $(V, E)$ and function $f$, $i$ is also a dummy node of the interaction-restricted function $f|_E$.

Consider any restricted null node $i$, we have 
\[
    f(T \cup i) = \sum_{R \in \zeta(T)} f(R) + f(i),
\]
for any connected subset $T \cup i \subseteq V$.

If $T\cup i$ is connected, we directly have $f|_E(T \cup i) = f(T \cup i) = f|_E(T) + f(i)$.

If $T \cup i$ is not connected, 
\begin{align}\label{eq:dum}
    f|_E(T \cup i) &= \sum_{R \in \zeta(T \cup i)} f(R).
\end{align}
Because $i$ is a restricted null node, $i$ does not provide any profit when joining others to form a connected subgraph, hence, for a connected component $R \in \zeta(T \cup i)$ such that $R \ni i$,  we have 
\begin{align*}
    f(R) &= \sum_{W \in \zeta(R \setminus i)} f(W) + f(i) \\
    &= f|_E (R \setminus i) + f(i).
\end{align*}
Combining with~\eqref{eq:dum}, we have $f|_E(T \cup i) = f|_E(T) + f(i)$ in case $T \cup i$ is not connected.

We then deduce that $i$ is also a dummy node with respect to the interaction-restricted function $f|_E$, \ie,
\[
    f|_E(T \cup i) = f|_E(T) + f|_E(i),
\]
for any $T \subseteq V$.

By the dummy feature axiom (Axiom~\ref{axm:shi-d}), we have 
\[
    \MyI^k_S (f, E) = \ShI^k_S(f|_E) = \begin{dcases}
        f(S) & \text{if~} |S| = 1, \\
        0 & \text{if~} S \ni i.
    \end{dcases}
\]
\end{proof}

\begin{proposition} \label{prop:myti-ce}
    The Myerson-Taylor index satisfies the component efficiency axiom (\bCE).
\end{proposition}

\begin{proof}[Proof of Proposition~\ref{prop:myti-ce}]
For any graph $(V, E)$, a function $f$, and a connected component $C \in \zeta(V)$, we define a characteristic game $f^C$ such that 
\[
    f^C|_E (T) = \sum_{R \in \zeta(T \cap C)} f(R) \quad \forall T \subseteq V.
\]
Because $f^C|_E (T) = f^C|_E (T \cap C)$ for any $T \subseteq V$, $C$ will be the carrier (the grand coalition) of $f^C|_E$, which means any nodes not in $C$ are dummy nodes. By the dummy axiom (Axiom~\ref{axm:shi-d}), we have $\MyI^k_{S}(f^C, E) =  \ShI^k_S(f^C|_E) = 0$, for all $S \not\subseteq C$ such that $|S| \leq k$. 

On the other hand, we notice that every connected component of a subgraph induced by a subset $T$ is also connected in the graph $(V, E)$, we thus have
\begin{align*}
    f|_E(T) &= \sum_{R \in \zeta(T)} f(R) \\
    &= \sum_{C \in \zeta(V) } \sum_{R \in \zeta(T \cap C)} f(R) \\
    &= \sum_{C \in \zeta(V)} f^C|_E (T).
\end{align*}

By the linearity axiom (Axiom~\ref{axm:shi-l}), for any component $C \in \zeta(V)$, we have
\begin{align*}
    \sum_{S \subseteq C, |S| \leq k} \MyI^k_S (f, E) &= \sum_{S \subseteq C, |S| \leq k} \ShI^k_S (f|_E) \\
    &= \sum_{C^\prime \in \zeta(V)} \sum_{S \subseteq C, |S| \leq k} \ShI^k_S(f^{C^\prime}|_E) \\
    &= \sum_{S \subseteq C, |S| \leq k} \ShI^k_S(f^{C}|_E) \\
    &= f^C|_E (C) = f(C).
\end{align*}
The third equality follows as any node $i \in C$ is the dummy node in the game of other component $f^{C'}|_E$, $C' \neq C$; thus, $\ShI_S^k(f^{C'}|_E) = 0$, for any $S \subseteq C$. The last equality follows from the efficiency axiom of the Shapley-Taylor index~(Axiom~\ref{axm:shi-e}). This completes the proof.
\end{proof}

\begin{proposition} \label{prop:myti-cf}
    The Myerson-Taylor index satisfies the coalitional fairness axiom (\bCF).
\end{proposition}

\begin{proof}[Proof of Proposition~\ref{prop:myti-cf}]
For any graph $(V, E)$ and a connected coalition $T$, consider two functions $f_1, f_2$ such that $f_1(R) = f_2(R), \forall R \neq T$. We define a function $g = f_1 - f_2$ so that, for any subset $R \subseteq V$, we have
\[
    g(R) = \begin{dcases}
        f_1(R) - f_2(R) & \text{if}~ R = T,\\
        0 & \text{otherwise}.
    \end{dcases}
\]
Thus, $g = \beta u_T$ where $\beta = f_1(T) - f_2(T)$ and $u_T$ is the unanimity function defined on the subset $T$. For any $S \subseteq T, |S| \leq k$, we have $\MyI^k_{S} (g, E) = \beta \ShI^k_{S}(u_T|_E) = \beta \ShI^k_{S} (u_T).$ The last equation is in fact that $T$ is connected.

By the symmetry axiom (Axiom~\ref{axm:shi-s}), we have $\ShI^k_{S_1} (u_T) = \ShI^k_{S_2} (u_T)$ for any $S_1, S_2 \subseteq T$ such that $|S_1| = |S_2|$. We deduce $\MyI^k_{S_1}(g, E) = \MyI^k_{S_2} (g, E)$. The proof follows from the linearity of the Myerson-Taylor index.
\end{proof}



\begin{proposition} \label{prop:myti-id}
    The Myerson-Taylor interaction index satisfies the interaction distribution axiom (\bID).
\end{proposition}

\begin{proof}[Proof of Proposition~\ref{prop:myti-id}]
    Consider unanimity function $u_T$ and a graph $(V, E)$ in which $T \subseteq V$ is connected. 
    
    For any $S \subsetneq T$ such that $|S| < k$, we have
    \begin{align*}
        \MyI^k_S (u_T, E) &= \ShI^k_S (u_T|_E) = \delta_S u_T|_E (\emptyset)\\
        &= \sum_{W \subseteq S} (-1)^{|S|-|W|} u_T|_E (W)\\
        &= \sum_{W \subseteq S} (-1)^{|S|-|W|} \sum_{R \in \zeta(W)} u_T(R) \\
        &= 0.
    \end{align*}
    
    The last equality follows as $u_T(R) = 0, \forall R \subseteq S \subsetneq T$.
    Hence, $\MyI^k_S(u_T, E) = 0$ for any $S \subsetneq T, |S| < k$.
\end{proof}

We present several elementary results needed to show the uniqueness.

The following result extends the result of~\citet[Lemma 1]{selccuk2014axiomatization} to $k$-order interactions.

\begin{restatable}[\citet{selccuk2014axiomatization}]{lemma}{lemmredu} \label{lemm:redu}
    If an interaction index $\Ic^k$ satisfies \bL~and \bRNP~axioms then, for any graph $(V, E)$, a function $f$, and a subset $S \subseteq V, |S| \leq k$, we have $\Ic^k_S (f, E) = \Ic^k_S(f|_E, E)$.
\end{restatable}

\begin{proof}[Proof of Lemma~\ref{lemm:redu}]
    Consider the function $g = f - f|_E$. We have
    \[
        g(T) = \begin{dcases}
        0 & \text{if $T$ is connected}, \\
        f(T) - \sum_{R \in \zeta(T)}(R) & \text{otherwise}.
    \end{dcases}
    \]
    
    Notice that, by definition, $g(T \cup i) - \sum_{R \in \zeta(T)} g(R) = 0$ for any node $i$ and connected coalition $T \cup i$. Thus, every node is a restricted null player in the game with the characteristic function $g$ and, by \bRNP~axiom, must receive zero payoffs. That is, $\Ic^k_S (g, E) = 0$ for any subset $S$. By \bL~axiom, we deduce that $\Ic^k_S (f, E) = \Ic^k_S(f|_E, E)$.
\end{proof}

\begin{proposition} \label{prop:alloc-unmt}
    Let $\Ic^k$ be a $k$-order interaction index that satisfies five axioms, \bL, \bRNP, \bCF, \bID, and~\bCE. For any graph $(V, E)$ and a connected subset $T$, then
    \[
        \Ic^k_S(u_T, E) = \begin{dcases}
            1 & \text{if}~ S = T ~\text{and}~ |S| < k, \\
            0 & \text{if}~ S \neq T ~\text{and}~ |S| < k, \\
            \frac{1}{{|T| \choose k}} & \text{if}~ S \subseteq T ~\text{and}~ |S| = k, \\
            0 & \text{if}~ S \not\subseteq T ~\text{and}~ |S| = k.
        \end{dcases}
    \]
\end{proposition}

\begin{proof}[Proof of Proposition~\ref{prop:alloc-unmt}]
    The procedure is similar to~\citet{sundararajan2020shapley} in which the restricted null player (\bRNP) replaces the dummy feature (Axiom~\ref{axm:shi-d}), and the coalitional fairness (\bCF) plays the role of symmetry axiom (Axiom~\ref{axm:shi-s}).
    
    We notice that any node $i \notin T$ is a restricted null node of $u_T$ as $T$ is connected. Hence, by \bRNP, $\Ic^k_S(u_T, E) = 0$ for any $S \setminus T \neq \emptyset$ (or $S \not\subseteq T$). 
    
    Consider the case $|T| < k$, for any subset $S$ such that $S \subsetneq T$, $\Ic^k_S(u_T, E) = 0$ by \bID~axiom. Therefore, $\Ic^k_S(u_T, E) = 0$ for any $S \neq T$. By \bCE~axiom, we have $\sum_S \Ic^k_S(u_T, E) = u_T(C) = 1$ where $C \in \zeta(V), C \supseteq T$ is a connected component containing $T$. As $\Ic^k_S(u_T, E) = 0$ for all $S \neq T$. We deduce $\Ic^k_S(u_T, E) = 1$ for $S = T$.

    For the case $|T| \geq k$, by \bRNP~and \bID~axiom, we also have $\Ic^k_S(u_T, E) = 0$ for any $S$ such that $S \not\subseteq T$ or $S \subsetneq T, |S| < k$. Hence, by \bCE~axiom, we have $\sum_{S \subseteq T, |S| = k} \Ic^k_S(u_T, E) = 1$. By the coalitional fairness axiom (\bCF), we then have $\Ic^k_S (u_T, E) = \frac{1}{{|T| \choose k}}$.

    This completes the proof.
\end{proof}

\begin{proof}[Proof of Theorem~\ref{thm:mt_uni}] 
    Propositions~\ref{prop:myti-l}-\ref{prop:myti-id} assert that the Myerson-Taylor index satisfies five axioms. The following shows its uniqueness.

    Let $\Ic^k$ be a $k$-order interaction index that satisfies five axioms. Consider a graph $(V, E)$ and a function $f \in \mc F^V$. By Lemma~\ref{lemm:redu}, we have $\Ic^k (f, E) = \Ic^k(f|_E, E)$ as $\Ic^k$ satisfies \bL~and \bRNP. Thus, we only need to show that $\mc I^k$ is a unique allocation rule in the space of interaction-restricted functions $\mc F^V|_E = \{f|_E : f \in \mc F^V\}$
    
     By~\citet[Lemma 2]{hamiache1999value}, an interaction-restricted function $f|_E \in \mc F^V|_E$ can be decomposed into
    \begin{align*}
        f|_E &= \sum_{\substack{T \subseteq V \\ T~\text{is connected}}} \Delta_{f|_E} (T) u_T, \\ 
        \text{where} \quad \Delta_{f|_E}(T) &= \sum_{\substack{R \subseteq T \subseteq \mc N(R)\\ R~\text{is connected}}} (-1)^{|T|-|R|} f(R).
    \end{align*}
    Here, $\mc N(R)$ is the set of vertices that are adjacent to $R$, \ie, $\mc N(R) = \{i : \exists ij \in E, j \in R\}$. Hence, the set of unanimity functions of connected subgraphs $\{u_T : T \subseteq V ~\text{where}~ T~\text{is connected} \}$ forms a basis of the space of $\mc F^V|_E$. By the \bL~axiom, we have
    \begin{align*}
        \Ic^k_S(f|_E, E) &= \sum_{\substack{T \subseteq V\\ T~\text{is connected}}} \Delta_{f|_E} (T) \Ic^k_S(u_T, E).
    \end{align*}
    Since $\Ic^k$ satisfies five axioms, $\Ic^k_S(u_T, E)$ is uniquely determined for any connected coalition $T$ by Proposition~\ref{prop:alloc-unmt}. Thus, $\Ic^k$ is uniquely determined for any interaction-restricted function $f|_E$.    
    This completes the proof. 
\end{proof}


    


    

\section{Additional Results}\label{sec:additional_results}
\subsection{Fairness of the Myerson-Taylor index}
We extend the Myerson value to high-order interaction using four axioms as in~\citep{selccuk2014axiomatization} as they align with the analyses for the Shapley-Taylor index~\citep{sundararajan2020shapley}. In what follows, we adapt the classical fairness axiom~\citep{myerson1977graphs} to higher-order interactions and show that the Myerson-Taylor index complies with this extended fairness criterion.

The following axiom extends the classical fairness axiom of the Myerson value~\citep{myerson1977graphs}.
\begin{property}[Fairness - \bf F]
    A $k$-order interaction index $\Ic^k$ is fair  if, for any graph $(V, E)$, characteristic function $f$, and $ij \in E$, we have
    \begin{equation*}
        \Ic^k_{S \cup i} (f, E) - \Ic^k_{S \cup i}(f, E \setminus {ij})
        = \Ic^k_{S \cup j}(f, E) - \Ic^k_{S \cup j}(f, E \setminus {ij}),
    \end{equation*}
    for all $S \subseteq V \setminus ij$ such that $|S| \leq k - 1$.
\end{property}


\begin{proposition} \label{prop:myti-f}
    The Myerson interaction index satisfies the fairness properties.
\end{proposition}


Before going to the proof of Proposition~\ref{prop:myti-f}, we present a generalized property from the equal treatment of equals in~\citep{beal2020necessary} to interaction indices.
\begin{property}[Equal treatment of equals]
    For any function $f$ any two equal nodes $i, j$ such that $f(T \cup i) = f(T \cup j), \forall T \subseteq V \setminus ij$, we then have $\Ic^k_{S \cup i} (f) = \Ic^k_{S \cup j} (f), \forall S \subseteq V \setminus ij, |S| \leq k - 1$.
\end{property}

It is known that equal treatment of equals is weaker than symmetry: any allocation rule satisfying symmetry also satisfies equal treatment of equals, while the reverse does not necessarily apply~\citep{beal2020necessary}. 

\begin{lemma}\label{lemm:eq_treat}
    Symmetry implies equal treatment of equals.
\end{lemma}

\begin{proof}
Consider a function $f$ and two equal nodes $i, j$. Let $\pi$ be a permutation of $V$ such that 
\[
    \pi(l) = \begin{dcases}
        i & \text{if}~ l = j, \\ 
        j & \text{if}~ l = i, \\
        l & \text{otherwise}.
    \end{dcases}
\]
We have $\pi f = f$ as $f(T \cup i) = f(T \cup j)$ for all $T \subseteq V \setminus ij$. By the symmetry axiom, we have $\mc I^k_{S \cup i} (f) = \mc I^k_{\pi (S \cup i)} (\pi f) = \mc I^k_{S \cup j} (f)$, for any $S \subseteq V \setminus ij$.
\end{proof}

We are now ready to prove Proposition~\ref{prop:myti-f}. In what follows, we use $\zeta(T, E\setminus ij)$ to denote a set of connected components of the subgraph of $(V, E\setminus ij)$ induced by a subset $T$.

\begin{proof}[Proof of Proposition~\ref{prop:myti-f}]
Consider any graph $(V, E)$, a function $f$, and any edge $ij \in E$, we define a characteristic function $g = f|_E - f|_{E \setminus ij}$. For any subset $T \subseteq V$, we have 
\[
    g(T) = \sum_{R \in \zeta(T, E)} f(R) - \sum_{R \in \zeta(T, E \setminus ij)} f(R).
\]

Thus $g(T) = 0$ if $\{i, j\} \not\subseteq T$ as removing the edge $ij$ does not affect the components in $T$. Consequently, for any $T \subseteq V \setminus ij$, $g(T \cup i) = g(T \cup j) = 0$.

In other words, $i$ and $j$ are equals in the game $g$. According to the equal treatment of equals property (Lemma~\ref{lemm:eq_treat}), we have $\ShI^k_{S \cup i}(g) = \ShI^k_{S \cup j}(g), \forall S \subseteq V \setminus ij, |S| \leq k - 1$. 

Using the linearity axiom and replacing the Shapley-Taylor index with the Myerson interaction index, we deduce
\[
    \MyI^k_{S \cup i}(f, E) - \MyI^k_{S \cup i}(f, E \setminus ij) = \MyI^k_{S \cup j} (f, E) - \MyI^k_{S \cup j} (f, E \setminus ij).
\]
This completes the proof.
\end{proof}

\subsection{Reduction}
The following result reduces the high-order interaction Shapley-Taylor into the classical Shapley value.
\begin{proposition}[Reduction] \label{prop:reduction}
    Let $\ShI^k$ is the $k$-order Shapley-Taylor interaction and $\Sh$ is the Shapley value, for any function $f$ and a player $i$, we have
    \[
        \phi_i (f) = \sum_{\substack{S \subseteq V \setminus i \\ |S \cup i| \leq k}} \frac{1}{|S \cup i|} \ShI^k_{S \cup i} (f).
    \]
\end{proposition}
\begin{proof}[Proof of Proposition~\ref{prop:reduction}]
    We show Propsition~\ref{prop:reduction} for any unanimity functions $u_T$, $T \subseteq V$. The proof for a general function $f$ follows by applying linear axiom (\bL) to the Shapley value and the Shapley-Taylor interaction index.
    
    We first consider the case that $i \notin T$, thus, by the dummy axiom (\bD), we have $\Sh_i(u_T) = \ShI^k_{S \cup i}(u_T) = 0$, for any $S \subseteq V \setminus i$.
    
    If $i \in T$, by the dummy axiom, we have
    \begin{align*}
        \sum_{\substack{S \subseteq V \setminus i \\ |S \cup i| \leq k}} \frac{1}{|S \cup i|} \ShI^k_{S \cup i} (u_T) &= \sum_{\substack{S \subseteq T \setminus i \\ |S \cup i| = k}} \frac{1}{k} \ShI^k_{S \cup i} (u_T) \\
        &= \sum_{\substack{S \subseteq T \setminus i \\ |S \cup i| = k}} \frac{1}{k} \frac{1}{{t \choose k}} \\
        &= \frac{1}{k} \frac{{t-1 \choose k-1}}{{t \choose k}} = \frac{1}{t}.
    \end{align*}
    The first equation follows by the \bID~axiom. The second equation follows by~\citet[Proposition 4]{sundararajan2020shapley}.

    It is known that $\Sh_i(T) = \frac{1}{t}, \forall i \in T$~\citep{algaba2019handbook}. We deduce that $\phi_i (f) = \sum_{\substack{S \subseteq V \setminus i \\ |S \cup i| \leq k}} \frac{1}{|S \cup i|} \ShI^k_{S \cup i} (f)$, for $i \in T$. 

    This completes the proof.
\end{proof}
Proposition~\ref{prop:reduction} indicates that the Shapley-Taylor index distributes the importance score of $i$, which is the Shapley value of $i$, to interactions up to size $k$ that node $i$ can join.

Since the Myerson value is a generalization of the Shapley value and the Myerson-Taylor index is a generalization of the Shapley-Taylor index, the reduction in Proposition~\ref{prop:reduction} holds for the Myerson values and the Myerson-Taylor index.

\section{Implementation Details} \label{sec:implementation}
\subsection{Myerson-Taylor Interaction Index}
We present a permutation-based sampling algorithm to compute the Myerson-Taylor index for a given graph input and black-box model in Algorithm~\ref{alg:myt}. The algorithm leverages the principle that the Shapley-Taylor index represents the expected value of discrete derivatives over a randomly chosen ordering of nodes in $V$. The main difference compared to the Shapley-Taylor is that the Myerson-Taylor uses the interaction-restricted function $f|_E$ (Line 8), instead of the vanilla $f$. Detailed implementation to compute the interaction-restricted value for a coalition $T$ is provided in Algorithm~\ref{alg:comres}.

\begin{figure*}[htb]
  \centering
  \begin{minipage}{.8\linewidth}
    \begin{algorithm}[H]
      \SetKwInOut{Param}{Param}
    \SetKwComment{Comment}{/* }{ */}
    \SetKwInOut{Input}{Input}
    \SetKwInOut{Output}{Output}
    \caption{Permutation-based sampling algorithm for the $k$-order Myerson-Taylor index.}
    \label{alg:myt}
    \Input{a graph $G = (V, E)$, a value function $f: 2^{|V|} \to \R$, order $k$}
    \Output{interaction index $\ve B$}
    $\ve A \gets \ve 0$ \Comment*[r]{Accumulated interactions}
    $\ve C \gets \ve 0$ \Comment*[r]{Count interactions}
    \For{$t = 0, 1, \ldots$}{
        $\pi \gets$ a random ordering of $\{1, 2, \ldots, |V|\}$\;
        \For{all subset $S \subseteq V$ with size $l$, $|S| = l$}{
            $i \gets$ left most index of $S$'s elements in the ordering $\pi$\;
            $T \gets \{\pi_1, \ldots, \pi_{i-1}\}$ \Comment*[r]{A set of predecessors of $S$ in $\pi$}
            $\ve A_S \gets \ve A_S + \delta_S f|_E(T)$ \Comment*[r]{$f|_E(T)$ is computed by Algorithm~\ref{alg:comres}}
            $\ve C_S \gets \ve C_S + 1$\;
        }
    }
    $\ve B \gets \ve 0$\;
    \For{every subset $S \subseteq V$ up to size $k$, $|S| \leq k$}{
        \eIf{$|S| \leq k$}{
             $\ve B_S \gets \delta_S f|_E(\emptyset)$\;
        }{
            $\ve B_S \gets \ve A_S / \ve C_S$\;
        }
    }
    \end{algorithm}
  \end{minipage}
\end{figure*}

\begin{figure*}[htb]
  \centering
  \begin{minipage}{.8\linewidth}
    \begin{algorithm}[H]
    
    \SetKwInOut{Param}{Param}
    \SetKwComment{Comment}{/* }{ */}
    \SetKwInOut{Input}{Input}
    \SetKwInOut{Output}{Output}
    \caption{Value of an interaction-restricted function ($f|_E(T)$).}\label{alg:comres}
    \Input{a graph $G = (V, E)$, a value function $f: 2^{|V|} \to \R$, a coalition $T$}
    \Output{A value $v \in \R$}
    $\zeta(T) \gets$ a set of components of subgraph $(T, E_T)$\;
    $v \gets 0$\;
    \For{every component $C$ in $\zeta(T)$}{
        $v \gets v + f(C)$\;
    }
\end{algorithm}
  \end{minipage}
\end{figure*}

\subsection{Motif Search}

We provide a linear relaxation approach to solve the problem~\eqref{eq:reform}, which can be explicitly rewritten as follows
\begin{align}
    \max_{S_1, \ldots, S_m \subseteq V} &~\rlap{$\displaystyle \sum_{l=1}^m \left| \sum_{\substack{i, j \in S_l \\ i \leq j}} \tau \ve B^+_{ij} + (1 - \tau) \ve B^-_{ij}\right|$} \\
    \st &~ S_l \cap S_h = \emptyset &1 \leq l, h \leq m \\
    &~ \left| \cup_{l=1}^m S_l \right| \leq M \\
    &~ (S_l, E_{S_l}) \text{~is connected} &1 \leq l \leq m \label{cons:cnvt}
\end{align}

The above optimization problem is a variant of the quadratic multiple knapsack problem~\citep{hiley2006quadratic} with absolute values. 
One strategy to solve~\eqref{eq:reform} is by linear relaxations and then using off-the-shelf MILP solvers such as MOSEK~\citep{mosek} or GUROBI~\citep{gurobi}.
The absolute operator in the objective can be cast to linear constraints using the big-M method, a widely used technique in integer programming to linearize constraints with absolute values. To capture the connectivity constraints~\eqref{cons:cnvt}, we employ the \textit{linear connectivity constraints} as in~\citep{mak2019mixed, althaus2014algorithms}. The linear connectivity constraints are initially relaxed and added in a \textit{lazy} fashion whenever the incumbent optimal solution violates them. Problem~\eqref{eq:reform} also exhibits symmetry in the solution (shuffling the set indices returns the same solution), and we break this symmetry using aggressive symmetry-breaking constraints of the MILP solvers.

\section{Experiments}\label{sec:expt}

\subsection{Experimental Details}\label{sec:exp_detail}

\subsubsection{Datasets}\label{sec:app_data}
We use the popular datasets in the literature of GNN explanability~\citep{sato2020survey, yuan2022explainability, agarwal2023evaluating}. 

\begin{itemize}
    \item \textit{BA-2Motifs~\citep{luo2020parameterized}}: The dataset is a binary classification task where each graph incorporates a Barabasi-Albert base structure linked with either a house or five-cycle motif. The label and ground-truth explanation of a graph is determined by the motif the graph contains.
    \item \textit{SPMotif~\citep{wu2022discovering}}: The dataset contains graphs combining a base structure (Tree, Ladder, or Wheel) and a motif (Cycle, House, Crane). A spurious correlation between the base and the motif is manually injected into each graph. A graph's label and ground truth explanation is determined based on the motif it contains.
    \item \textit{BA-HouseGrid}: Similar to BA-2Motifs but with two distinct motifs, house and $3\times 3$ grid. The house and grid motif are chosen because they do not have overlapping structures such as those found in the house and five-cycle.
    \item \textit{BA-HouseAndGrid}: Each graph is a Barabási structure that may linked with either house or grid motifs. Graphs containing both motif types are labeled as 1, otherwise 0. 
    \item \textit{BA-HouseOrGrid}: Similar to BA-HouseAndGrid, however, graphs with either house or grid motifs are labeled as 1, otherwise 0.
    \item \textit{Mutagenic~\citep{kazius2005derivation}}: The dataset is a molecular property prediction task, which is to identify if a molecule is mutagenic or not. The functional groups -NO2 and -NH2 are considered as ground-truth explanations that lead to mutagenicity~\citep{luo2020parameterized}.
    \item \textit{Benzene~\citep{sanchez2020evaluating}}: The dataset contains 12000 molecular graphs extracted from ZINC15~\citep{sterling2015zinc}. The task is to determine the presence of benzene rings in a molecule. Ground-truth explanations are the carbon atoms in the benzene rings.
    \item \textit{MNIST75SP~\citep{monti2017geometric}}: An image classification dataset where each image in MNIST is converted to a superpixel graph. Each node represents a superpixel where node features are the superpixel's central coordinate and brightness intensity. The spatial proximity between the superpixels determines the edges. Graph-truth explanations are the top 15 superpixels with the highest intensity.
    \item \textit{GraphSST2 and Twitter~\citep{yuan2022explainability}}: The datasets are sentiment classification tasks. GraphSST2 has two classes, and Twitter has three classes. Each node corresponds to one word in the text, edges are constructed by the Biaffine parser~\citep{gardner2018allennlp}, and node features are pre-trained BERT embedding of words. No ground-truth explanations are available for these datasets.
\end{itemize}
Table~\ref{tab:datastats} provides statistics for chosen datasets. 

\begin{table}[t]
\centering
\caption{Statistics and properties of datasets. The datasets above the dashed blue line are synthetic, and below the dashed blue line are real-world ones.}
\label{tab:datastats}
\scriptsize
\begin{filecontents*}{data_stats.csv}
Dataset, no. graphs,no. classes,no. nodes,no. edges
BA-2Motifs,1000,2,25.00,25.48
BA-HouseGrid,10000,2,26.97,28.95
SPMotif,18000,3,45.98,66.72
BA-HouseAndGrid,10000,2,29.42,41.01
BA-HouseOrGrid,10000,2,24.73,34.31
Mutagenic,2951,2,30.13,30.45
Benzene,12000,2,20.58,21.83
MNIST75SP,70000,10,70.57,295.25
GraphSST2,70042,2,10.20,9.20
Twitter,6940,3,21.10,20.10
\end{filecontents*}
\pgfplotstabletypeset[
    col sep=comma,
    string type,
    every head row/.style={before row=\toprule},
    every head row/.style={before row=\toprule, after row=\midrule},
    every last row/.style={after row=\bottomrule},
    every row no 4/.style={after row=\tabucline[0.4pt blue!40 off 2pt]{-}},
]{data_stats.csv}
\end{table}

\subsubsection{Models} \label{sec:app_model}
In this paper, we use three GNN models commonly used in explainability literature: GCN~\citep{kipf2016semi}, GIN~\citep{xu2018powerful}, and GAT~\citep{velivckovic2017graph}. The accuracy of these models is provided in Table~\ref{tab:modelacc}. As SubgraphX released the checkpoint for GCN for BA-2Motifs, we used their checkpoint. As GAT performs poorly on synthetic datasets, we only explain for GAT on real-world datasets. Note that GAT's poor performance on synthetic datasets is also reported in~\citep[Table 3]{amara2023ginx} and~\citep[Table 2]{zheng2024ci}.

\begin{table}[t]
\centering
\caption{Accuracy of GNNs models on evaluated datasets. The datasets above the dashed blue line are synthetic, and below the dashed blue line are real-world ones.}
\label{tab:modelacc}
\scriptsize
\begin{filecontents*}{model_acc.csv}
Dataset,GCN,GIN,GAT
BA-2Motifs,97.00,100.00,41.00
BA-HouseGrid,100.00,99.90,52.30
SPMotif,86.00,96.17,50.11
BA-HouseAndGrid,99.30,99.90,51.10
BA-HouseOrGrid,100.00,100.00,51.10
Mutagenic,89.86,91.55,92.57
Benzene,85.42,91.92,85.92
MNIST75SP,82.44,89.43,94.64
GraphSST2,89.13,87.86,89.24
Twitter,69.22,63.87,67.34
\end{filecontents*}
\pgfplotstabletypeset[
    col sep=comma,
    string type,
    every head row/.style={before row=\toprule},
    every head row/.style={before row=\toprule, after row=\midrule},
    every last row/.style={after row=\bottomrule},
    every row no 4/.style={after row=\tabucline[0.4pt blue!40 off 2pt]{-}},
]{model_acc.csv}
\end{table}

\begin{table*}[t]
\centering
\caption{Explanation accuracy for the GIN model on datasets with a single motif. Note that GradCAM is a gradient-based method; GNNExplainer, PGExplainer, Refine, and MatchExplainer are perturbation-based; The last four methods are cooperative game-based.}
\label{tab:single_motif_gin}
\scriptsize
\begin{filecontents*}{single_motif_gin_no_std.csv}
methods,BA-2Motifs,BA-2Motifs,BA-HouseGrid,BA-HouseGrid,SPMotif,SPMotif,MNIST75SP,MNIST75SP
,\textit{F1} $\uparrow$,\textit{AUC}$\uparrow$,\textit{F1} $\uparrow$,\textit{AUC}$\uparrow$,\textit{F1} $\uparrow$,\textit{AUC}$\uparrow$,\textit{F1} $\uparrow$,\textit{AUC}$\uparrow$
GradCAM,\cellcolor{green!25}\textbf{0.934},\cellcolor{green!25}\textbf{0.988},\cellcolor{yellow!25}\underline{0.876},\cellcolor{green!25}\textbf{0.978},\cellcolor{green!25}\textbf{0.776},\cellcolor{green!25}\textbf{0.887},0.230,0.486
GNNExplainer,0.280,0.486,0.360,0.508,0.176,0.491,0.257,0.525
PGExplainer,0.314,0.363,0.449,0.491,0.429,0.415,0.202,0.484
Refine,0.116,0.427,0.231,0.506,0.153,0.475,0.194,\cellcolor{yellow!25}\underline{0.543}
MatchExplainer,0.672,0.765,0.482,0.635,0.189,0.494,0.196,0.499
SubgraphX,0.863,0.901,0.862,0.839,0.484,0.673,0.205,0.501
GStarX,0.185,0.484,0.268,0.544,0.170,0.486,\cellcolor{yellow!25}\underline{0.307},0.532
SAME,\cellcolor{yellow!25}\underline{0.909},0.914,0.846,0.827,0.347,0.607,0.275,0.525
MAGE (ours),\cellcolor{green!25}\textbf{0.940},\cellcolor{yellow!25}\underline{0.942},\cellcolor{green!25}\textbf{0.954},\cellcolor{yellow!25}\underline{0.942},\cellcolor{yellow!25}\underline{0.630},\cellcolor{yellow!25}\underline{0.729},\cellcolor{green!25}\textbf{0.586},\cellcolor{green!25}\textbf{0.675}
\end{filecontents*}
\pgfplotstabletypeset[
    col sep=comma,
    string type,
    every head row/.style={before row=\toprule},
    every row no 0/.style={after row=\midrule},
    every head row/.style={output empty row, before row={%
            \toprule \multirow{2}{*}{Method}  &
            \multicolumn{2}{c}{BA-2Motifs} & \multicolumn{2}{c}{BA-HouseGrid} & \multicolumn{2}{c}{SPMotif} & \multicolumn{2}{c}{MNIST75SP} \\
            \cmidrule(r){2-3} \cmidrule(r){4-5} \cmidrule(r){6-7} \cmidrule(r){8-9}
        }},
    every row no 1/.style={after row=\tabucline[0.4pt blue!40 off 2pt]{-}},
    every row no 5/.style={after row=\tabucline[0.4pt blue!40 off 2pt]{-}},
    every last row/.style={after row=\bottomrule},
]{single_motif_gin_no_std.csv}
\end{table*}

\begin{table*}[t]
\centering
\caption{Explanation accuracy for the GCN model on datasets with multiple motifs.}
\label{tab:multi_motifs_gcn}
\scriptsize
\begin{filecontents*}{multi_motifs_gcn_no_std.csv}
methods,BA-HouseAndGrid,BA-HouseAndGrid,BA-HouseOrGrid,BA-HouseOrGrid,Mutagenic,Mutagenic,Benzene,Benzene
,\textit{AMI}$\uparrow$,\textit{AUC}$\uparrow$,\textit{AMI}$\uparrow$,\textit{AUC}$\uparrow$,\textit{AMI}$\uparrow$,\textit{AUC}$\uparrow$,\textit{AMI}$\uparrow$,\textit{AUC}$\uparrow$
GradCAM,0.570,0.776,\cellcolor{yellow!25}\underline{0.980},\cellcolor{green!25}\textbf{0.998},0.423,0.887,\cellcolor{green!25}\textbf{0.669},\cellcolor{green!25}\textbf{0.928}
GNNExplainer,0.242,0.491,0.184,0.485,0.151,0.578,0.173,0.492
PGExplainer,0.483,0.699,0.428,0.541,0.689,\cellcolor{yellow!25}\underline{0.949},0.216,0.056
Refine,0.241,0.457,0.114,0.452,0.463,0.876,0.199,0.618
MatchExplainer,\cellcolor{yellow!25}\underline{0.612},\cellcolor{yellow!25}\underline{0.842},0.606,\cellcolor{yellow!25}\underline{0.818},0.231,0.601,0.165,0.508
SubgraphX,0.532,0.714,0.533,0.779,0.672,0.840,0.326,0.643
GStarX,0.201,0.497,0.130,0.481,0.003,0.472,0.049,0.515
SAME,0.528,0.714,0.631,\cellcolor{yellow!25}\underline{0.813},\cellcolor{yellow!25}\underline{0.787},0.895,0.109,0.536
MAGE (ours),\cellcolor{green!25}\textbf{0.675},\cellcolor{green!25}\textbf{0.879},\cellcolor{green!25}\textbf{0.999},\cellcolor{green!25}\textbf{0.999},\cellcolor{green!25}\textbf{0.987},\cellcolor{green!25}\textbf{0.993},\cellcolor{yellow!25}\underline{0.584},\cellcolor{yellow!25}\underline{0.784}
\end{filecontents*}
\pgfplotstabletypeset[
    col sep=comma,
    string type,
    every head row/.style={before row=\toprule},
    every row no 0/.style={after row=\midrule},
    every head row/.style={output empty row, before row={%
            \toprule \multirow{2}{*}{Method}  &
            \multicolumn{2}{c}{BA-HouseAndGrid} & \multicolumn{2}{c}{BA-HouseOrGrid} & \multicolumn{2}{c}{Mutagenic} & \multicolumn{2}{c}{Benzene} \\
            \cmidrule(r){2-3} \cmidrule(r){4-5} \cmidrule(r){6-7} \cmidrule(r){8-9}
        }},
    every row no 1/.style={after row=\tabucline[0.4pt blue!40 off 2pt]{-}},
    every row no 5/.style={after row=\tabucline[0.4pt blue!40 off 2pt]{-}},
    every last row/.style={after row=\bottomrule},
]{multi_motifs_gcn_no_std.csv}
\end{table*}

\begin{table*}[t]
\caption{Results on sentiment classification tasks. Note that GradCAM is a gradient-based method, while the other methods are cooperative game-based.}\label{tab:sst_gin_gat}
\begin{subtable}[h]{0.47\linewidth}
\centering
\caption{GIN}
\scriptsize
\begin{filecontents*}{sst_gin_no_std.csv}
methods,GraphSST2,GraphSST2,Twitter,Twitter
,$\mathrm{Fid}_{\alpha}$ $\uparrow$,$\mathrm{Fid}$ $\uparrow$,$\mathrm{Fid}_{\alpha}$ $\uparrow$,$\mathrm{Fid}$ $\uparrow$
GradCAM,\cellcolor{yellow!25}\underline{0.140},0.259,0.251,0.412
SubgraphX,\cellcolor{yellow!25}\underline{0.147},\cellcolor{green!25}\textbf{0.307},0.239,0.398
GStarX,0.132,0.264,0.243,0.403
SAME,\cellcolor{yellow!25}\underline{0.149},\cellcolor{yellow!25}\underline{0.281},\cellcolor{yellow!25}\underline{0.301},\cellcolor{yellow!25}\underline{0.475}
SAGE (ours),\cellcolor{green!25}\textbf{0.172},\cellcolor{green!25}\textbf{0.309},\cellcolor{green!25}\textbf{0.374},\cellcolor{green!25}\textbf{0.590}
\end{filecontents*}
\pgfplotstabletypeset[
    col sep=comma,
    string type,
    every head row/.style={before row=\toprule},
    every row no 0/.style={after row=\midrule},
    every head row/.style={output empty row, before row={%
            \toprule \multirow{2}{*}{Method}  &
            \multicolumn{2}{c}{GraphSST2} & \multicolumn{2}{c}{Twitter} \\
            \cmidrule(r){2-3} \cmidrule(r){4-5}
        }},
    every row no 1/.style={after row=\tabucline[0.4pt blue!40 off 2pt]{-}},
    every last row/.style={after row=\bottomrule},
]{sst_gin_no_std.csv}
\end{subtable}
\begin{subtable}[h]{0.47\linewidth}
\centering
\caption{GAT}
\label{tab:sst_gat}
\scriptsize
\begin{filecontents*}{sst_gat_no_std.csv}
methods,GraphSST2,GraphSST2,Twitter,Twitter
,$\mathrm{Fid}_{\alpha}$ $\uparrow$,$\mathrm{Fid}$ $\uparrow$,$\mathrm{Fid}_{\alpha}$ $\uparrow$,$\mathrm{Fid}$ $\uparrow$
GradCAM,--,--,--,--
SubgraphX,0.112,0.192,\cellcolor{yellow!25}\underline{0.086},0.129
GStarX,\cellcolor{green!25}\textbf{0.141},\cellcolor{yellow!25}\underline{0.250},\cellcolor{yellow!25}\underline{0.093},0.147
SAME,\cellcolor{yellow!25}\underline{0.123},0.203,\cellcolor{green!25}\textbf{0.106},\cellcolor{yellow!25}\underline{0.158}
SAGE (ours),\cellcolor{green!25}\textbf{0.148},\cellcolor{green!25}\textbf{0.266},\cellcolor{green!25}\textbf{0.112},\cellcolor{green!25}\textbf{0.176}
\end{filecontents*}
\pgfplotstabletypeset[
    col sep=comma,
    string type,
    every head row/.style={before row=\toprule},
    every row no 0/.style={after row=\midrule},
    every head row/.style={output empty row, before row={%
            \toprule \multirow{2}{*}{Method}  &
            \multicolumn{2}{c}{GraphSST2} & \multicolumn{2}{c}{Twitter} \\
            \cmidrule(r){2-3} \cmidrule(r){4-5} 
        }},
    every row no 1/.style={after row=\tabucline[0.4pt blue!40 off 2pt]{-}},
    every last row/.style={after row=\bottomrule},
]{sst_gat_no_std.csv}
\end{subtable}
\end{table*}

\subsubsection{Metrics} \label{sec:app_metric}

For a GNN model $f$ and input graph $G = (V, E)$, the output of explainers is a subgraph $(S, E_S)$. Let $(S^{gt}, E^{gt})$ be the ground truth explanation. Following previous practice in~\citep{ying2019gnnexplainer, luo2020parameterized}, we use the below metrics for datasets with ground truth explanations:
\begin{itemize}
    \item \textit{Area Under the ROC Curve (AUC)}: To measure AUC metrics, we treat the explanation task as the binary classification task on the edges of the input graph. We then compute the explanation accuracy by comparing binary edge masks generated by explainers $E_S$ against the ground truth edge masks corresponding to $E^{gt}$.
    \item \textit{F1 score}: The F1 score reports the overlap of the nodes highlighted by the explainers $S$ compared to the ground truth explanation $S^{gt}$. We use the F1 score for datasets where graphs contain only a single motif.
    \item \textit{Adjusted Mutual Information (AMI)}: AMI score is a common metric for evaluating different clustering algorithms. We use the AMI score to measure explainers' ability to identify different explanatory substructures. Specifically, for an explanation $S = (S_1, \ldots, S_m)$ and ground truth explanation $S^{gt} = (S^{gt}_1, \ldots S^{gt}_l)$, the mutual information between two partitions $S$ and $S^{gt}$ is calculated by
    \[
        \mathrm{MI}(S, S^{gt}) = \sum_{i = 1}^m \sum_{j = 1}^l p(i, j) \log \frac{p(i, j)}{p(i) p'(j)},
    \]
    where $p(i) = \frac{|S_i|}{|V|}$ is the probability that a node picked at random from explanation $S$ falls into a motif $S_i$. Similarity, we have $p'(j) = \frac{|S^{gt}_j|}{|V|}$ and $p(i, j) = \frac{|S_i \cap S^{gt}_j|}{|V|}$. The vanilla MI tends to favor partitions with a higher number of clusters, regardless of the actual amount of `mutual information' between the label assignments, \textit{adjusted} MI (AMI) is used to alleviate this bias~\citep{vinh2009information}. Even though most of the current explainers do not consider the multiple motif setting, they might highlight multiple disconnected components. For these baselines, we consider each connected component in the highlighted subgraph as one identified motif.
\end{itemize}

For datasets without ground truth explanations, we use the following metrics:
\begin{itemize}
    \item \textit{Fidelity ($\mathrm{Fid}$)~\citep{pope2019explainability, yuan2021explainability}}: Fidelity measures the faithfulness of an explanation by reporting the changes in the model output when removing or keeping only selected nodes $S$
    \begin{align*}
        \mathrm{Fid}^{+} (S) &= f(V) - f(V \setminus S), \\
        \mathrm{Fid}^{-} (S) &= f(V) - f(S), \\ 
        \mathrm{Fid} (S) &= \mathrm{Fid}^{+} (S) - \mathrm{Fid}^{-} (S).
    \end{align*}
    The higher fidelity means that $S$ is more important to the model prediction.
    \item \textit{Robust Fidelity} ($\mathrm{Fid_{\alpha}}$): Since it is known that $\mathrm{Fid}$ is sensitive to the OOD samples, we report $\mathrm{Fid}_\alpha$, a new metric proposed in~\citep{zheng2023towards} to alleviate the OOD problem of $\mathrm{Fid}$
    \begin{align*}
        \mathrm{Fid}^{+}_\alpha (S) &= f(V) - \mathbb{E} f(V \setminus \Omega^\alpha(S) ), \\
        \mathrm{Fid}^{-}_\alpha (S) &= f(V) - \mathbb{E} f(S \cup \Omega^{1-\alpha} (V \setminus S)), \\ 
        \mathrm{Fid}_\alpha (S) &= \mathrm{Fid}^{+}_\alpha (S) - \mathrm{Fid}^{-}_\alpha (S),
    \end{align*}
    where $\Omega^\alpha(T), 0 \leq \alpha \leq 1$ is a random subset of $T$ where a node in $T$ is included with probability $\alpha$ and erased with probability $1 - \alpha$. The key idea of $\mathrm{Fid}_\alpha$ is that if a subset $S$ is important to the model prediction, removing or keeping a superset of $S$ should also change the model output significantly. In case $\alpha =  1$, $\mathrm{Fid}_\alpha$ coincides with the original $\mathrm{Fid}$ metric. In this paper, we set $\alpha = 0.8$. 
\end{itemize}

\subsubsection{Experimental Setup and Implementation}
We mainly follow the experimental settings as in~\citep{yuan2021explainability, yuan2022explainability}, where we leverage their codebase\footnote{https://github.com/divelab/DIG/tree/dig-stable/benchmarks/xgraph} for GNN models and baseline explainers. We use the default hyperparameters for the baselines as in~\citep{yuan2022explainability}.

Following~\citep{zheng2023towards}, we only explain for well-trained models with reasonable performance and graph instances that the GNN models correctly predict. We split the dataset into training, validation, and test subsets with respective ratios of 0.8, 0.1, and 0.1. We train GNN models to a reasonable performance and then run the explainers for graph instances in the test datasets. We report the average metrics over instances in the test dataset. For synthetic and molecular datasets, we explain all instances in the test set with ground truth explanations. For MNIST75SP, GraphSST2, and Twitter, we randomly select 200 instances for evaluation. 

\subsection{Remaining Results}\label{sec:app_result}

\subsubsection{Quantitative Results}
We provide the remaining results for other combinations of GNN models and datasets in Table~\ref{tab:single_motif_gin},~\ref{tab:multi_motifs_gcn}, and~\ref{tab:sst_gin_gat}. In terms of explanation accuracy (Table~\ref{tab:single_motif_gin} and Table~\ref{tab:multi_motifs_gcn}), MAGE outperforms other cooperative game-based explainers in most settings while providing competitive performance with GradCAM, a while-box explainer.


\subsubsection{Ablation Study}

\noindent\textbf{Ablation for the group attribution.} This experiment validates that higher-order interactions can better approximate the group attribution than node-wise values such as Shapley and Myerson values.

For a given group $S$, the Shapley ($\phi$) or Myerson ($\psi$) values estimate the contribution of a group $S$ to the model prediction by the sum of node-wise importance
\[
    \mathrm{GrAttr}(\psi, S) = \sum_{i \in S} \psi_i.
\]
The change in the model prediction in the absence of the group $S$ is then estimated by
\begin{align*}
    \mathrm{GrAttr}(\psi, \bar{S}) &= \mathrm{GrAttr}(\psi, V) - \mathrm{GrAttr}(\psi, V \setminus S) \\
    &= \mathrm{GrAttr}(\psi, S).
\end{align*}
Here, we can see that $\mathrm{GrAttr}(\psi, \bar{S}) = \mathrm{GrAttr}(\psi, S)$ when we use node-wise importance to approximate the group attribution.

Meanwhile, we can estimate the group attribution of $S$ using the second-order Shapley-Taylor ($\Phi$) or Myerson-Taylor ($\Psi$) index as follows
\begin{align*}
    \mathrm{GrAttr}(\Psi, S) &= \sum_{i, j \in S} \Psi_{ij} \\ 
    \mathrm{GrAttr}(\Psi, \bar{S}) &= \mathrm{GrAttr}(\Psi, V) - \mathrm{GrAttr}(\Psi, S).
\end{align*}

To show that second-order interaction indices provide a better approximation of the group attribution, we conduct an experiment on image classification tasks where the Shapley values were more frequently used. Specifically, we use ResNet50~\citep{he2016deep} and ViT-B/16~\citep{dosovitskiy2020image} models pre-trained on Imagenet. We then compute four attribution methods, including the Shapley values, Myerson values, $2^{nd}$-order Shapley-Taylor index, and $2^{nd}$-order Myerson-Taylor index for 50 representative images provided by SHAP~\citep{lundberg2017unified}. Each image is segmented into 49 ($7 \times 7$) patches to compute the Shapley values and Shapley-Taylor indices. We build a grid graph on 49 patches as the interaction-restricted function for the Myerson value and Myerson-Taylor index. Two patches are connected if they are spatially adjacent.

We measure the faithfulness of the group attributions of four methods using infidelity metric~\citep{yeh2019fidelity}, which evaluates the difference between the estimated group attribution and the model's prediction in the presence ($\mathrm{Infid^+}$) and absence of the group ($\mathrm{Infid^-}$)
\begin{align*}
    \mathrm{Infid^+}(\mc I^k) &= \mathbb E_S \left| \mathrm{GrAttr}(\mc I^k, S) - (f(S) - f(\emptyset)) \right|, \\
    \mathrm{Infid^-}(\mc I^k) &= \mathbb E_S \big| \mathrm{GrAttr}(\mc I^k, \bar{S}) - [f(V) - f(V \setminus S)] \big|.
\end{align*}
The lower infidelity indicates that the group attribution method more accurately captures potential shifts in the model's predictions in the presence or absence of a group.

\begin{table*}[h!]
    \centering
    \caption{Ablation study for computing group attribution on Imagenet. The last column is the number of model queries needed to compute the attributions.}\label{tab:grattr}
    \scriptsize
    \begin{tabular}{cccccc}
        \toprule
        \multirow{2}{*}{Method} & \multicolumn{2}{c}{ResNet50} & \multicolumn{2}{c}{ViT-B/16} & \\
        \cmidrule(r){2-3} \cmidrule(r){4-5}
        & $\mathrm{Infid}_{+} (\downarrow)$ & $\mathrm{Infid}_{-} (\downarrow)$  & $\mathrm{Infid}_{+} (\downarrow)$ & $\mathrm{Infid}_{-} (\downarrow)$ &\#queries \\
        \midrule
        Shapley ($\phi$) & 0.083 & 0.123 & 0.058 & \cellcolor{yellow!25}\underline{0.065} & 18K \\ 
        Myerson ($\psi$) & 0.083 & 0.123 & \cellcolor{yellow!25}\underline{0.057} & \cellcolor{green!25}\textbf{0.064} & 13K \\ 
        Shapley-Taylor ($\ShI^2$) & \cellcolor{yellow!25}\underline{0.060} & \cellcolor{yellow!25}\underline{0.110} & 0.062 & 0.105 & 57K \\
        Myerson-Taylor ($\MyI^2$) & \cellcolor{green!25}\textbf{0.056} & \cellcolor{green!25}\textbf{0.108} & \cellcolor{green!25}\textbf{0.032} & 0.075 & 19K \\
        \bottomrule
    \end{tabular}
\end{table*}

Table~\ref{tab:grattr} shows that second-order indices (Shapley-Taylor and Myerson-Taylor) provide a better approximation for group attributions than using (first-order) marginal contributions (Shapley and Myerson). Notably, Myerson-based attributions improve the estimation of group attributions with fewer queries on the black-box model. 


\noindent\textbf{Ablating the interaction indices by edge-based explainers.} To investigate the effectiveness of the interaction indices compared to simple edge-based explainers that assign an importance score for every edge in the graph, we ablate the interaction indices with edge-based explainers in our MAGE framework. Here, the edge-based explainer could act as the interaction matrix $\ve B$ where the interactions between two nodes without a direct connection will be zero. We apply our motif search component directly upon this interaction matrix to find explanatory motifs.

We conduct experiments with two edge-based explainers: GNNExplainer and EdgeShaper, and their combinations with our motif search module. GNNExplainer~\cite{ying2019gnnexplainer} is not a game-based method, so it may not have theoretical properties like Shapley values. Therefore, we also apply EdgeShaper~\cite{mastropietro2022edgeshaper}, which applies Shapley values on edges (consider an edge as a player) to compute importance scores for edges. EdgeShaper is an edge-wise importance method based on Shapley value, thus has Shapley’s properties. However, it does not satisfy the interaction distribution (ID) axiom, which supports the Shapley-Taylor and Myerson-Taylor indices.

The result shown in table~\ref{tab:edge-based-motif-search} shows the enhanced performance of edge-based explanations when augmented with the motif search module in most of the settings. However, this is not always the case since we can also observe a drop from EdgeShaper + Motif Search compared to EdgeShaper. This drop may be because EdgeShaper does not align with the definition of the group attribution in the motif search module since EdgeShaper (and other edge-based explainers) assume that two nodes without a direct edge do interact with each other and will have a zero interaction score. However, Myerson-Taylor may attribute a positive (or negative) interaction score to connected nodes from multiple hops away. Thanks to that, our framework (Myerson-Taylor + Motif Search) outperforms other ablated methods by a considerable margin.

\begin{table*}[ht!]
\centering
\caption{Performance of MAGE when ablating the interaction indices by edge-based explainers.}
\label{tab:edge-based-motif-search}
\scriptsize
\begin{filecontents*}{edge_based_motif_search.csv}
,Ba2motif,Ba2motif,BaHouseGrid,BaHouseGrid,Mutag0,Mutag0,Benzene,Benzene,GraphSST2,GraphSST2
,F1,AUC,F1,AUC,AMI,AUC,AMI,AUC,$\mathrm{Fid}_\alpha$,$\mathrm{Fid}$
GNNExplainer,0.22,0.44,0.30,0.55,0.23,0.68,0.18,0.49,-0.02,0.07
GNNExplainer + Motif Search,0.43,0.64,0.60,0.75,0.06,0.46,0.27,0.51,0.08,0.15
EdgeShaper,0.48,0.72,0.49,0.65,0.70,0.86,0.37,0.70,-0.11,0.03
EdgeShaper + Motif Search,0.55,0.71,0.67,0.77,0.59,0.79,0.40,0.71,-0.13,0.01
Myerson-Taylor + Motif search (\textbf{MAGE}),\textbf{0.86},\textbf{0.89},\textbf{0.83},\textbf{0.85},\textbf{1.00},\textbf{1.00},\textbf{0.92},\textbf{0.96},\textbf{0.20},\textbf{0.34}
\end{filecontents*}
\pgfplotstabletypeset[
    col sep=comma,
    string type,
    every head row/.style={before row=\toprule},
    every row no 0/.style={after row=\midrule},
    every head row/.style={output empty row, before row={%
            \toprule \multirow{2}{*}{Method}  &
            \multicolumn{2}{c}{BA-2Motifs} & \multicolumn{2}{c}{BA-HouseGrid} & \multicolumn{2}{c}{Mutagenic} & \multicolumn{2}{c}{Benzene} & \multicolumn{2}{c}{GraphSST2} \\
            \cmidrule(r){2-3} \cmidrule(r){4-5} \cmidrule(r){6-7} \cmidrule(r){8-9} \cmidrule(r){10-11}
        }},
    every last row/.style={after row=\bottomrule},
]{edge_based_motif_search.csv}
\end{table*}

\noindent\textbf{Ablation on the higher-order interaction indices.} In the main paper, we use the second-order interaction index mainly because of its efficacy and efficiency. Using a higher-order interaction index ($k > 2$) will require more computation cost in approximating the interaction index and solving the motif search problem. For example, if we use $k = 3$, the motif search problem would become a cubic program, which would be more difficult to solve. For demonstration purposes, we conduct an ablation study using a third-order interaction index ($k=3$) and apply linear relaxation to solve the motif search. The result is reported for BA-2Motifs in Table~\ref{tab:higher-orders}.

\begin{table}[ht!]
\centering
\caption{Comparison of higher-order Shapley-Taylor and Myerson-Taylor.}
\label{tab:higher-orders}
\scriptsize
\begin{filecontents*}{higher_order_comparison.csv}
 ,F1,AUC,Queries,Running time (s)
MAGE (Shapley-Taylor $k = 2$),0.709,0.787,51K,11.4
MAGE (Myerson-Taylor $k = 2$),0.86,0.89,7K,5.1
MAGE (Shapley-Taylor $k = 3$),0.92,0.93,343K,104.2
MAGE (Myerson-Taylor $k = 3$),0.96,0.97,25K,56.8
\end{filecontents*}
\pgfplotstabletypeset[
    col sep=comma,
    string type,
    every head row/.style={before row=\toprule},
    every row no 0/.style={before row=\midrule},
    every head row/.style={output empty row, before row={%
            \toprule \multicolumn{1}{c}{ } & \multicolumn{1}{c}{F1} & \multicolumn{1}{c}{AUC} & \multicolumn{1}{c}{$\# Q$} & \multicolumn{1}{c}{Running Time (s)} \\
        }},
    every last row/.style={after row=\bottomrule},
]{higher_order_comparison.csv}
\end{table}

We observe that third-order interaction indices increase the performance on the BA-2Motifs dataset compared to second-order interaction indices. However, as a trade-off, the number of model queries and the running time increase significantly.

\subsubsection{Complexity Analysis}
\noindent\textbf{Running time analysis.} Table~\ref{tab:runtime} shows the running time for our method and competing methods on evaluated datasets. Note that due to the additional training stages or the necessity for access to the model's gradient or training data, GradCAM, PGExplainer, Refine, and MatchExplainer are excluded from the direct comparison. Therefore, our comparisons are focused on GNN explainers and other cooperative game-based methods. Significantly, MAGE exhibits the most competitive running time among cooperative game-based explainers while showing superior performance.

\begin{table*}[!h]
\centering
\caption{The average running time of competing methods on evaluated datasets (second/sample).}
\label{tab:runtime}
\scriptsize
\begin{filecontents*}{gcn_runtime.csv}
Method,BA-2Motifs,BA-HouseGrid,SPMotif,MNIST75SP,BA-HouseAndGrid,BA-HouseOrGrid,Mutagenic,Benzene
GNNExplainer,2.27,2.30,3.50,15.83,2.62,3.41,3.78,4.91
SubgraphX,66.60,54.40,74.14,823.55,70.18,81.41,63.63,4.40
GStarX,19.63,15.53,21.51,35.75,29.47,17.06,16.43,20.02
SAME,11.61,44.42,19.03,142.11,90.73,41.27,6.12,7.14
MAGE (ours),5.16,6.18,14.61,44.28,17.74,9.38,4.89,3.63
\end{filecontents*}
\pgfplotstabletypeset[
    col sep=comma,
    string type,
    every head row/.style={before row=\toprule},
    every head row/.style={before row=\toprule, after row=\midrule},
    every last row/.style={after row=\bottomrule},
]{gcn_runtime.csv}
\end{table*}

\begin{table*}[h!]
    \caption{Model query efficiency across game-based methods.} 
    \label{tab:queries}
    \resizebox{\textwidth}{!}{
    \begin{filecontents*}{queries.csv}
  ,Ba2motifs,BaHouseGrid,Spmotif,Mnist75sp,GraphSST2,Twitter,ba house and grid,ba house or grid,mutag0,benzene
,$\#Q \downarrow$,$\#Q \downarrow$,$\#Q \downarrow$,$\#Q \downarrow$,$\#Q \downarrow$,$\#Q \downarrow$,$\#Q \downarrow$,$\#Q \downarrow$,$\#Q \downarrow$,$\#Q \downarrow$
SubgraphX,333K,340K,375K,2865K,255K,313K,430K,350K,258K,19K
SAME,45K,232K,108K,881K,24K,31K,466K,217K,27K,28K
GStarX,26K,28K,37K,$\textbf{72K}$,18K,21K,34K,30K,28K,23K
MAGE,$\textbf{7K}$,$\textbf{9K}$,$\textbf{18K}$,128K,$\textbf{2K}$,$\textbf{3K}$,$\textbf{17K}$,$\textbf{15K}$,$\textbf{3K}$,$\textbf{1K}$
\end{filecontents*}
\pgfplotstabletypeset[
        col sep=comma,
        string type,
        every head row/.style={before row=\toprule},
        every row no 0/.style={after row=\midrule},
        every head row/.style={output empty row, before row={%
                \toprule \multirow{3}{*}{Method} &
                \multicolumn{4}{c}{Single Motif} & \multicolumn{2}{c}{Sentiment Analysis} &
                \multicolumn{4}{c}{Multiple Motifs}\\ 
                \cmidrule(r){2-5} \cmidrule(r){6-7} \cmidrule(r){8-11}
                &\multicolumn{1}{c}{BA-2Motifs} & 
                \multicolumn{1}{c}{BA-HouseGrid} & 
                \multicolumn{1}{c}{SPMotif} & 
                \multicolumn{1}{c}{MNIST75SP} & 
                \multicolumn{1}{c}{GraphSST2} & 
                \multicolumn{1}{c}{Twitter} & 
                \multicolumn{1}{c}{BA-HouseAndGrid} & \multicolumn{1}{c}{BA-HouseOrGrid} & \multicolumn{1}{c}{Mutagenic} & \multicolumn{1}{c}{Benzene} \\
            }},
        every last row/.style={after row=\bottomrule},      
    ]{queries.csv}
    }
\end{table*}

\noindent\textbf{Analysis of the number of model queries.} 
We report the average number of model queries needed to explain an instance by game-based explainers. Table~\ref{tab:queries} shows that MAGE requires fewer queries than other game-based methods, especially when compared with MCTS methods such as SubgraphX and SAME. MAGE demonstrates its superiority by reducing queries by averaged factors of 51.40 (SubgraphX), 14.63 (SAME), and 6.17 (GStarX).

\subsubsection{Qualitative Results}\label{sec:app_qualitative}

\begin{figure*}[h]
    \centering
    \includegraphics[width=\linewidth]{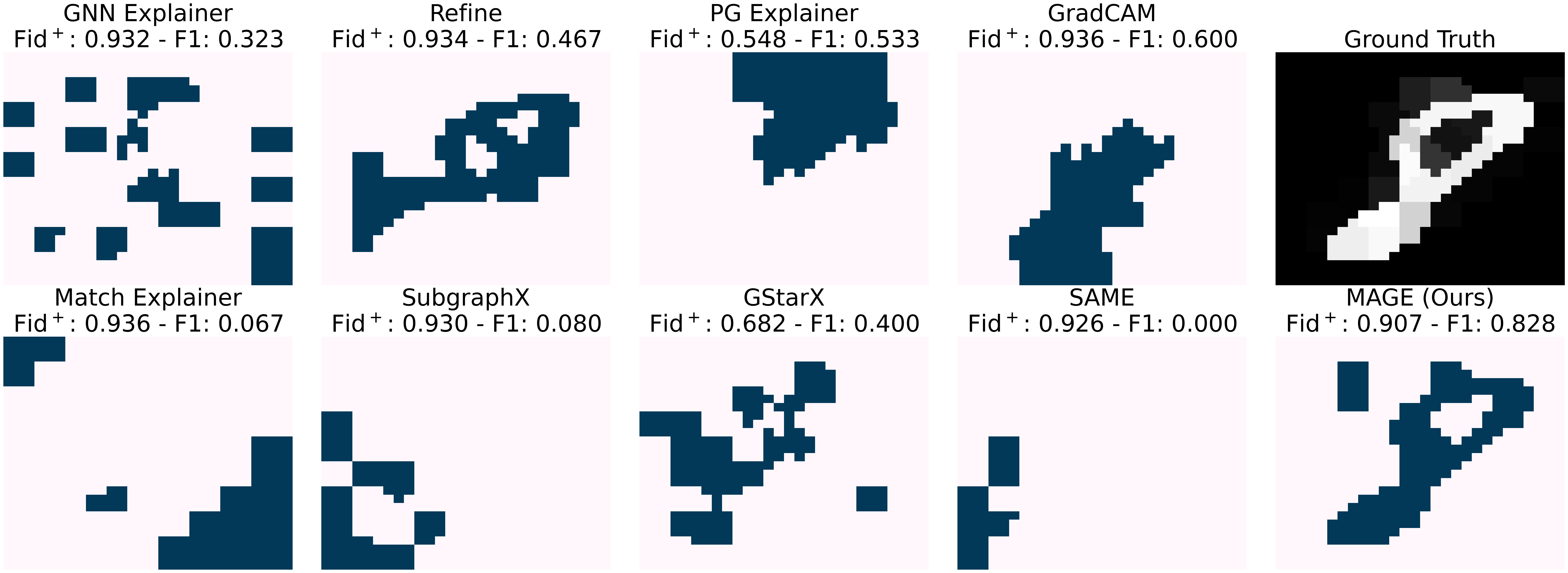}
    \caption{This example visualizes the explanation for the GCN model of MAGE against competing baselines on MNIST75SP. Despite achieving high fidelity ($\mathrm{Fid}^+$) scores, the explanations of baselines are not meaningful. Meanwhile, only MAGE can generate an explanation that aligns with pixels that describe number `8'}
    \label{fig:mnist_11453}
\end{figure*}

\begin{figure*}[h]
    \centering
    \begin{subfigure}{\linewidth}
        \centering
        \includegraphics[width=0.6\linewidth]{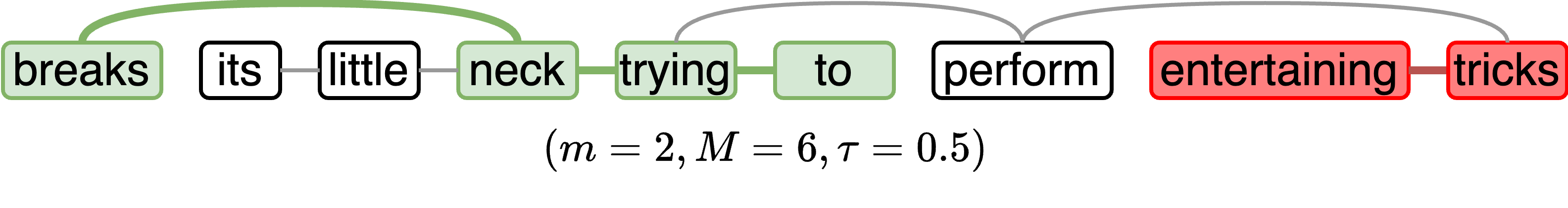}
        \caption{Model prediction: Negative sentiment. MAGE highlights the main negative verb phrase `breaks neck', which contributes to the overall negative sentiment, and the phrase `entertaining tricks', which shows a slightly positive sentiment.}
    \end{subfigure}
    \begin{subfigure}{\linewidth}
        \centering
        \includegraphics[width=\linewidth]{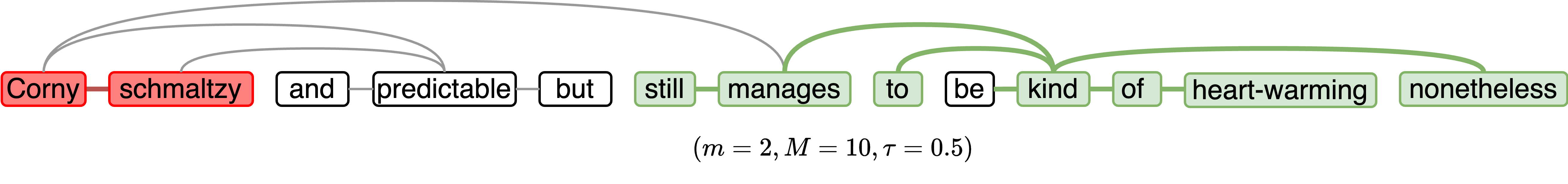}
        \caption{Model prediction: positive sentiment. MAGE highlights the main adjective, `heart-warming,' which contributes to the overall positive sentiment, and two minor adjectives, `corny' and `schmaltzy,' which display some negative sentiment.}
    \end{subfigure}
    \caption{MAGE can highlight text subgraphs with contradicting sentiments in GraphSST2 Dataset.} 
    \label{fig:graphsst2_neg}
\end{figure*}

\begin{figure*}
    \centering
    \includegraphics[width=\linewidth]{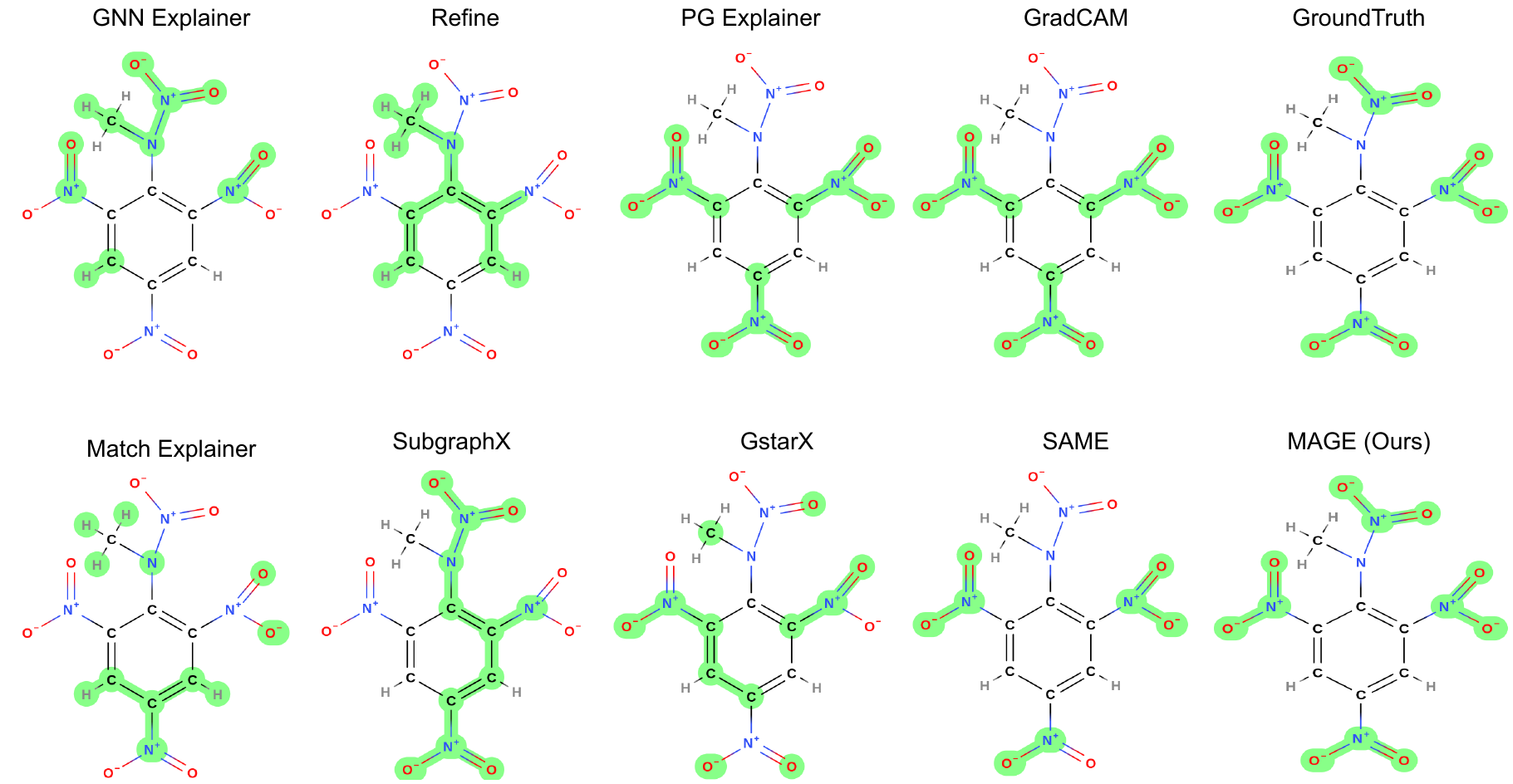}
    \caption{Explanations of competing methods on a molecgraph from Mutagenic dataset.}
    \label{fig:mutag_2913}
\end{figure*}

\begin{figure*}
    \centering
    \includegraphics[width=0.97\linewidth]{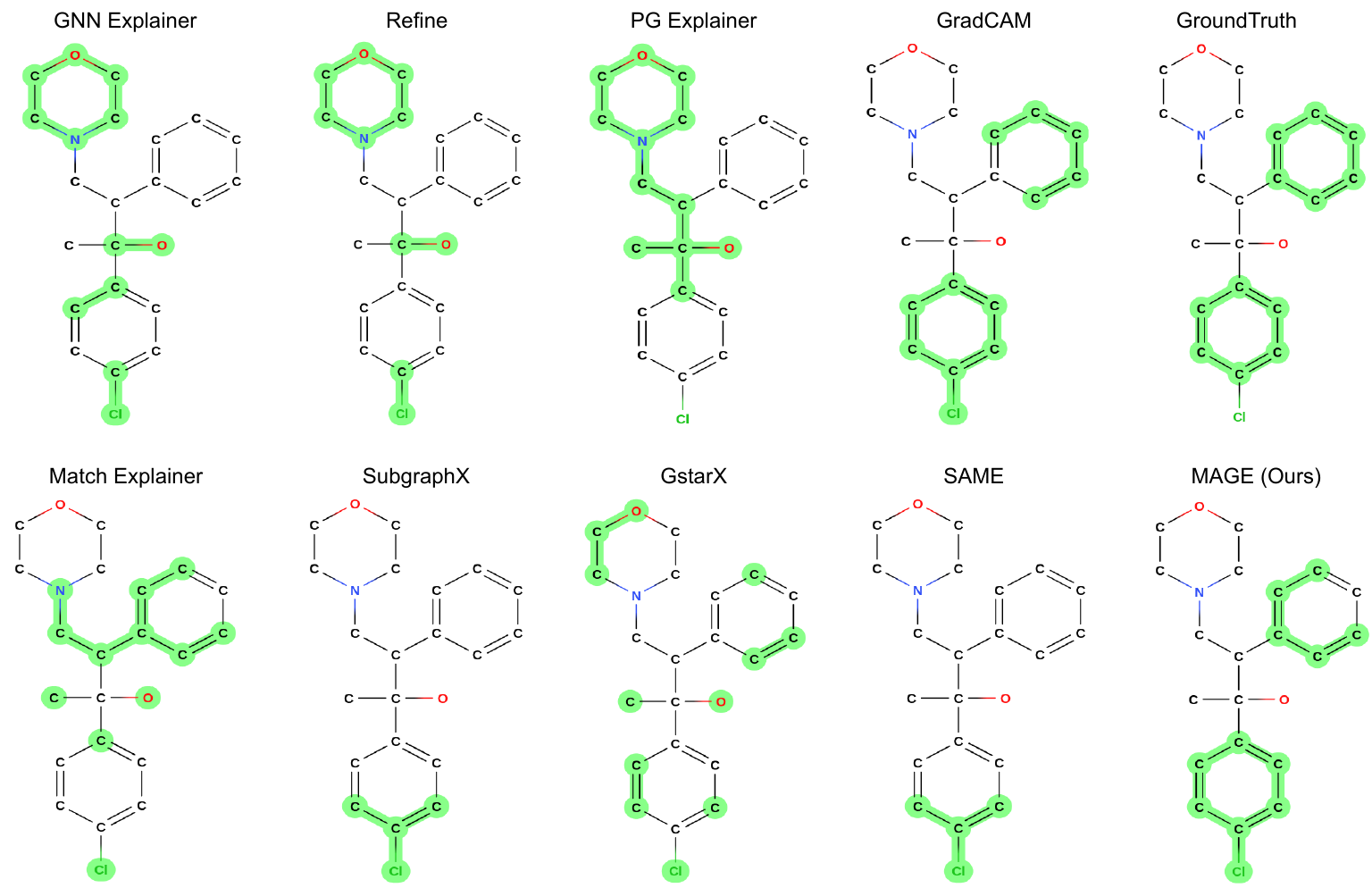}
    \caption{Explanations of competing methods on a molecular graph from Benzene dataset.}
    \label{fig:benzene_5898}
\end{figure*}

\begin{figure*}
    \centering
    \includegraphics[width=1.0\linewidth]{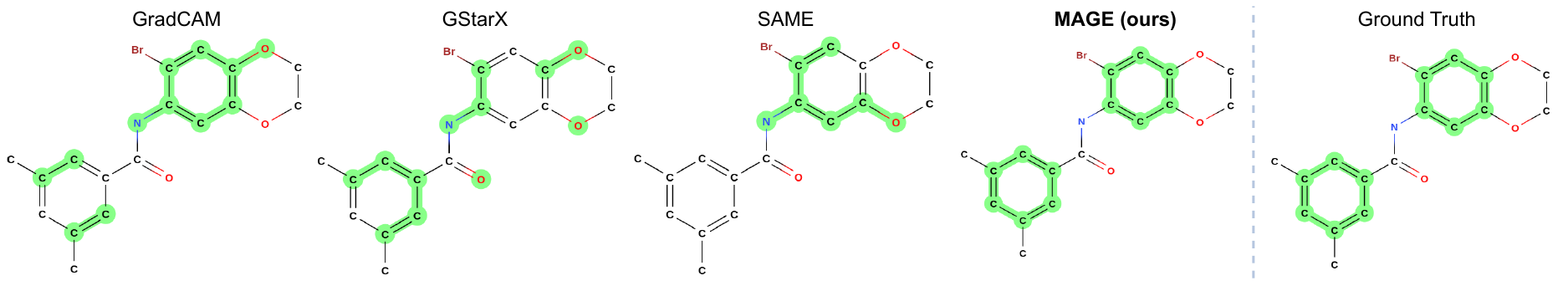}
    \captionof{figure}{Molecule C17H16BrNO3 input is predicted in class `have benzene ring' by GNN. Our MAGE multi-motif explanations correctly identify the two benzene rings; while competing methods such as GradCAM, GStarX, and SAME fail.} 
    \label{fig:intro_example}
\end{figure*}

\begin{figure*}
    \centering
    \includegraphics[width=\linewidth]{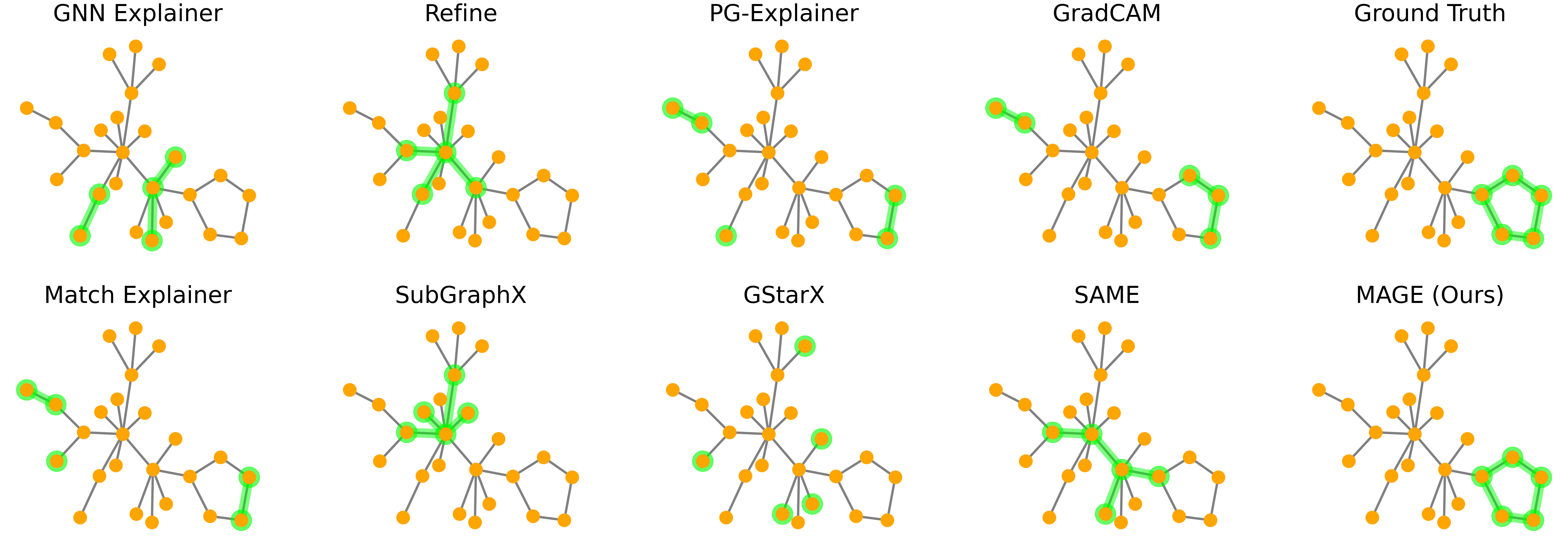}
    \caption{Explanations of competing methods on a synthetic graph from BA-2Motifs dataset.}
    \label{fig:ba_2motifs_495}
\end{figure*}

\begin{figure*}
    \centering
    \includegraphics[width=\linewidth]{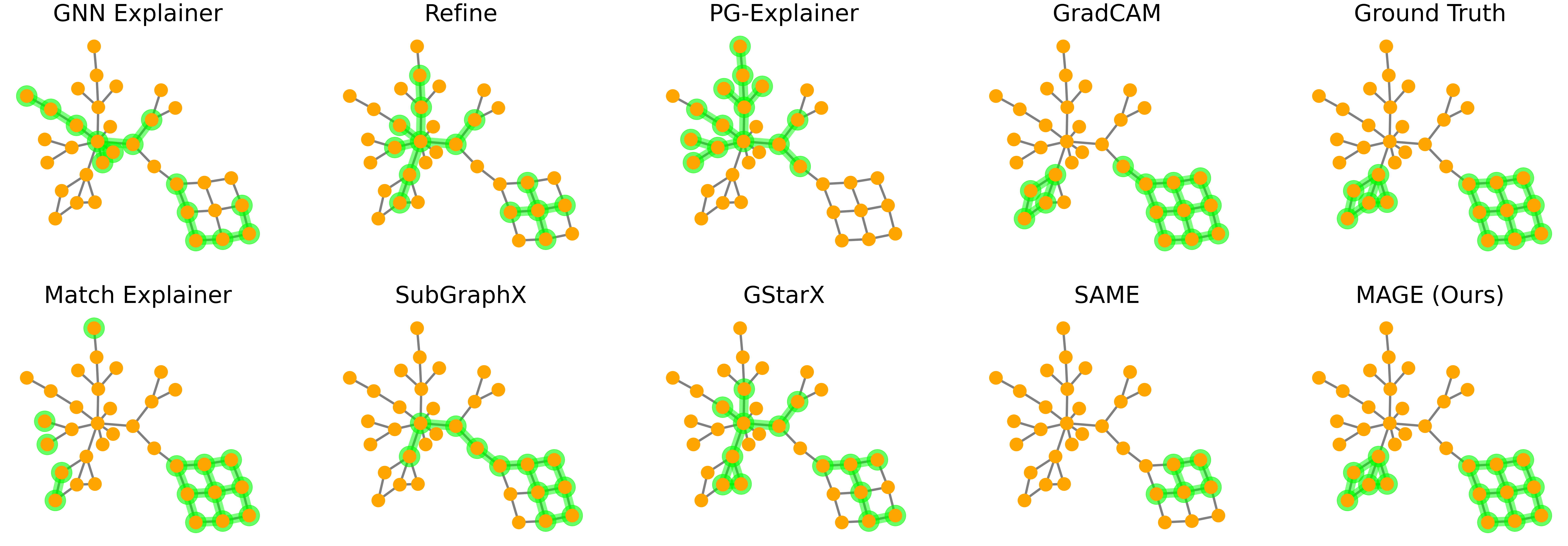}
    \caption{Explanations of competing methods on a synthetic graph from BA-HouseOrGrid dataset.}
    \label{fig:ba_house_or_grid_5120}
\end{figure*}

\begin{figure*}
    \centering
    \includegraphics[width=\linewidth]{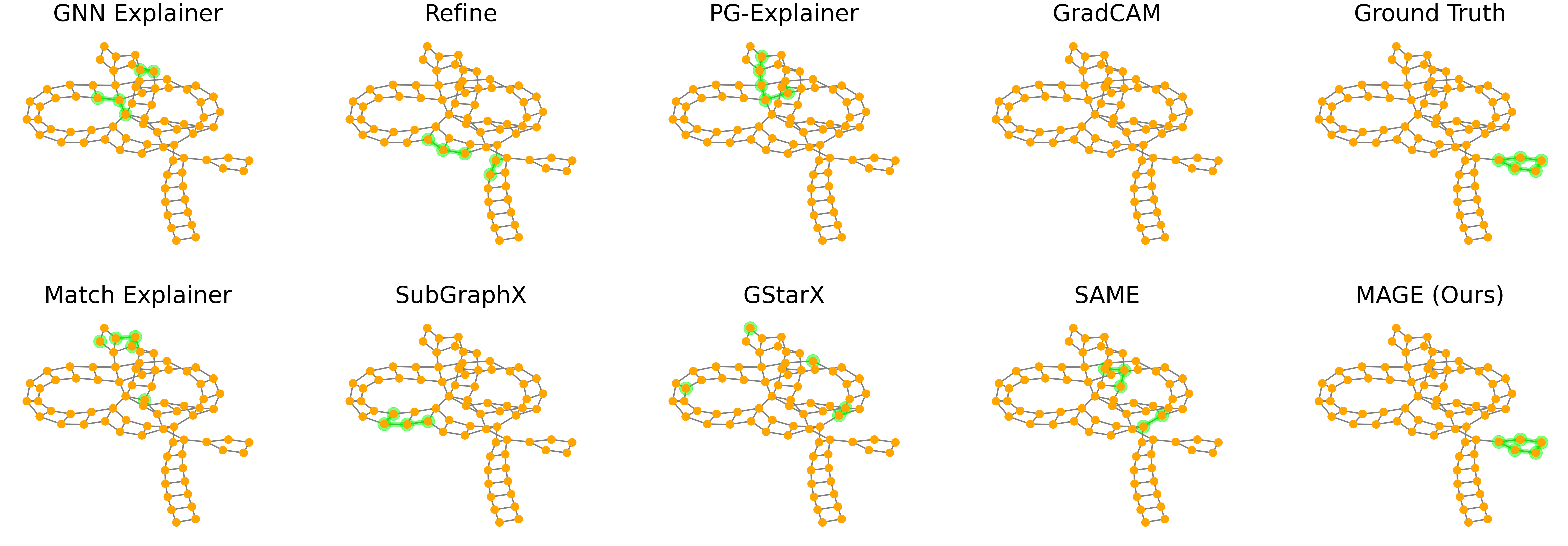}


\caption{Explanations of competing methods on a synthetic graph from SPMotif dataset.}
\label{fig:spmotif_11236}
\end{figure*}

For image classification (Figure~\ref{fig:mnist_11453}), only MAGE can provide a meaningful explanation that aligns with the brightest superpixels in the input image. Meanwhile, graph explainers fail to provide meaningful explanations for class `8', despite having high fidelity scores. This observation aligns with arguments in~\citep{zheng2023towards}, suggesting that the fidelity metric is sensitive to out-of-distribution explanations. Note that GradCAM~\citep{pope2019explainability} is adapted to explain GNN models on the graph inputs constructed from superpixels in the original images, which is different from typical GradCAM~\cite{selvaraju2017grad} running on images. We also provide more examples in the GraphSST2 dataset (Figure~\ref{fig:graphsst2_neg}). MAGE can highlight both structures that support and contradict the sentiment predicted by the model prediction. We provide more examples for Mutagentic, Benzene, and synthetic datasets in Figure~\ref{fig:mutag_2913}-\ref{fig:spmotif_11236}.

\end{document}